\PassOptionsToPackage{table}{xcolor}
\PassOptionsToPackage{pagebackref,breaklinks=true,colorlinks,citecolor=blue,urlcolor=blue,linkcolor=blue,bookmarks=false}{hyperref}
\PassOptionsToPackage{noabbrev,nameinlink,capitalize}{cleveref}

\documentclass[onecolumn,numbers]{april_aigc}

\usepackage[utf8]{inputenc}
\usepackage{url}
\usepackage{amsfonts}
\usepackage{amssymb}
\usepackage{nicefrac}
\usepackage{amsmath}
\usepackage{tabularx}
\usepackage{array}
\usepackage{enumitem}
\usepackage{adjustbox}
\usepackage{pifont}
\usepackage{afterpage}
\usepackage{wrapfig}
\usepackage{listings}
\usepackage{fancyhdr}
\tcbuselibrary{listings,breakable}

\titleformat*{\section}{\color{aprilblue}\Large\sffamily\bfseries}
\titleformat*{\subsection}{\color{aprilblue}\large\sffamily\bfseries}
\titleformat*{\subsubsection}{\color{aprilblue}\normalsize\sffamily\bfseries}

\lstdefinestyle{prompttemplate}{
  basicstyle=\small\ttfamily,
  breaklines=true,
  breakatwhitespace=false,
  columns=fullflexible,
  keepspaces=true,
  showstringspaces=false
}

\newtcblisting{prompttemplatebox}[1]{
  listing only,
  breakable,
  colback=black!2,
  colframe=black!70,
  coltitle=white,
  colbacktitle=black!75,
  fonttitle=\bfseries,
  title={#1},
  boxrule=0.6pt,
  arc=2pt,
  left=6pt,
  right=6pt,
  top=5pt,
  bottom=5pt,
  listing options={style=prompttemplate}
}

\newcounter{paperalgorithm}
\newenvironment{paperalgorithm}[1]{%
  \refstepcounter{paperalgorithm}%
  \par\noindent\begin{minipage}{\linewidth}
  \hrule\vspace{0.35ex}
  \noindent\textbf{Algorithm~\thepaperalgorithm\ #1}
  \vspace{0.35ex}\hrule\vspace{0.6ex}
}{%
  \vspace{0.4ex}\hrule
  \end{minipage}\par
}

\renewcommand{\title}[1]{\def\titlelist{{\fontsize{20pt}{28pt}\selectfont\sffamily\bfseries #1}}}
\title{Advancing Narrative Long Video Generation via Training-Free Identity-Aware Memory}

\author[1]{Jinzhuo Liu}
\author[1\raisebox{-0.2em}{\includegraphics[height=0.85em]{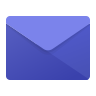}}]{Jiangning Zhang}
\author[1]{Wencan Jiang}
\author[1,2]{Yabiao Wang}
\author[3]{Dingkang Liang}
\author[1]{Zhucun Xue}
\author[4]{Ran Yi}
\author[1]{Yong Liu}
\affiliation[1]{Zhejiang University}
\affiliation[2]{Tencent Youtu Lab}
\affiliation[3]{Huazhong University of Science and Technology}
\affiliation[4]{Shanghai Jiao Tong University}
\coverdate{May 19, 2026}
\covercorrespondence{jinzhuo.liu@zju.edu.cn; 186368@zju.edu.cn}
\coversourcecode{https://github.com/Eddie0521/IAMFlow}
\coverproject{https://eddie0521.github.io/projects/iamflow}

\hypersetup{
  pdftitle={Advancing Narrative Long Video Generation via Training-Free Identity-Aware Memory},
  pdfauthor={Jinzhuo Liu, Jiangning Zhang, Wencan Jiang, Yabiao Wang, Dingkang Liang, Zhucun Xue, Ran Yi, Yong Liu},
  pdfsubject={Training-free identity-aware memory for narrative long video generation},
  pdfkeywords={IAMFlow, narrative video generation, autoregressive video generation, identity-aware memory, NarraStream-Bench}
}

\abstract{
Autoregressive video generation has improved rapidly in visual fidelity and interactivity, but it still suffers from long-term inconsistency and memory degradation.
Most existing solutions either compress historical frames using predefined strategies or retrieve keyframes based on coarse implicit attention signals, both of which fail to handle evolving prompts with shifting entity references, leading to identity drift, character duplication, and attribute loss.
To address this, we propose \textbf{IAMFlow}, a training-free identity-aware memory framework that explicitly models and tracks persistent entity identities, enabling consistent generation across prompt transitions.
Specifically, an LLM extracts entities with visual attributes from each prompt and assigns unique global IDs for identity-aware memory, while a VLM asynchronously verifies and refines attributes from rendered frames, enabling explicit entity tracking in place of implicit similarity-based matching.
To keep the proposed framework computationally practical, we design a systematic inference acceleration pipeline, including asynchronous visual verification, adaptive prompt transition, and model quantization, which achieves faster generation than existing baselines.
Furthermore, we introduce \textbf{NarraStream-Bench}, a benchmark for narrative streaming video generation that features 324 multi-prompt scripts spanning six dimensions and a three-dimensional evaluation protocol that integrates both traditional metrics and multimodal large language model-based assessments.
Extensive experiments show that IAMFlow, despite being training-free, achieves the best overall performance on NarraStream-Bench, outperforming the strongest baseline by \textbf{2.56} points, while achieving a \textbf{1.39$\times$} speedup over the most efficient baseline in the 60-second multi-prompt setting.
}

\begin{document}

\maketitle
\thispagestyle{plain}

\afterpage{%
\begin{figure}[t]
  \centering
  \includegraphics[width=\linewidth]{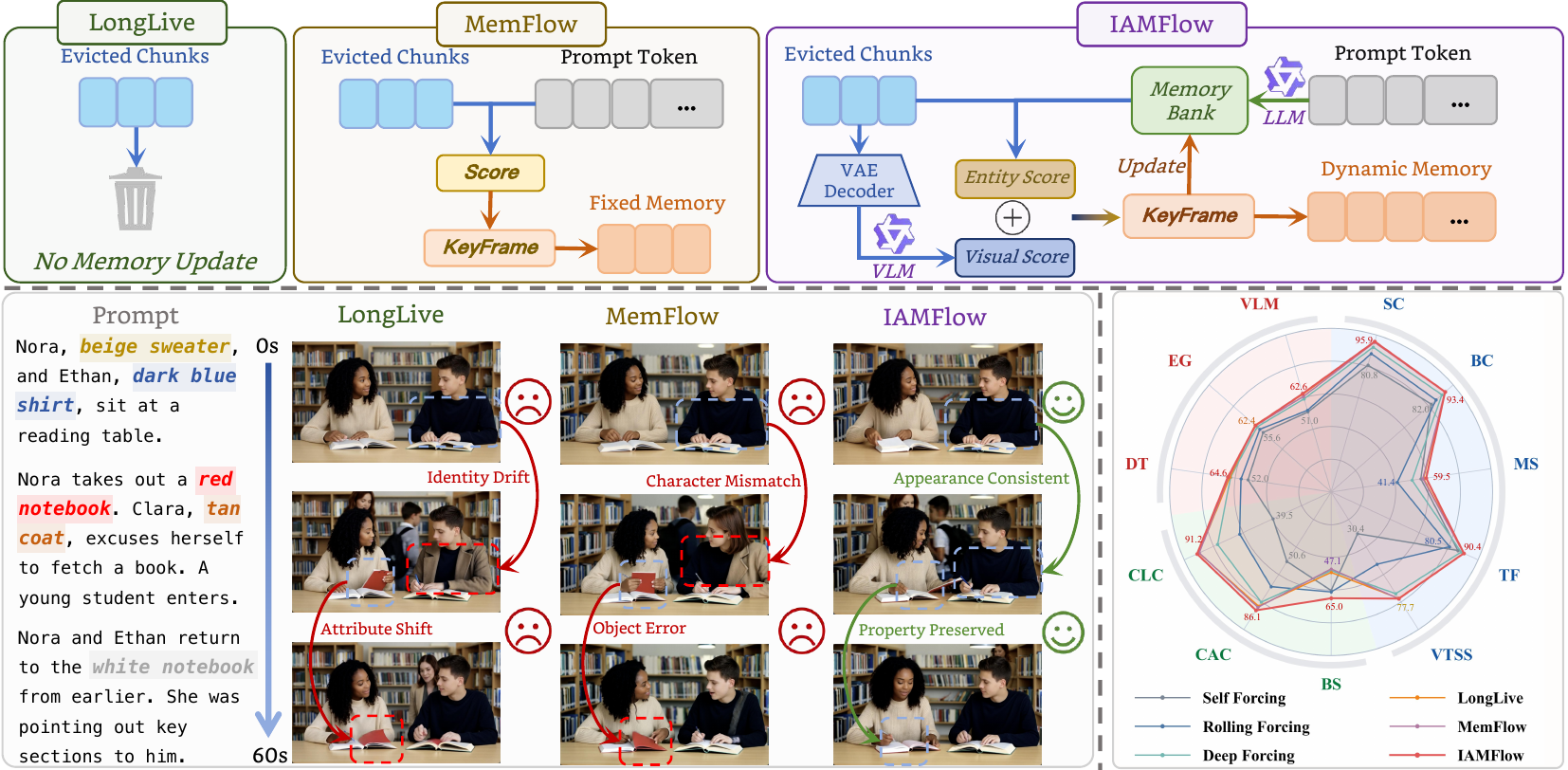}
  \caption{
  \textbf{Top}: Comparison of three representative memory paradigms.
  \textbf{Bottom}: Under evolving narrative prompts, LongLive and MemFlow fail to preserve long-term consistency, while IAMFlow preserves entity identities and attributes over 60 seconds and obtains superior performance across 11 metrics (see the radar chart).}
  \label{fig:fig1_teaser}
  \vspace{-5pt}
  \end{figure}
}

\section{Introduction}

Recent advances in video generation have been largely driven by Diffusion Transformer (DiT) models~\cite{dit,wan21,cogvideox} with bidirectional attention. However, the quadratic cost of full attention~\cite{attention} limits these models to generate short videos in a single pass, motivating autoregressive (AR) video generation~\cite{diffusionforcing,nova,causvid} as a more scalable paradigm for long and interactive video synthesis. Despite this promise, the AR paradigm still faces three core issues:
(i) \textit{limited historical memory}~\cite{longlive,memflow,framepack}, where causal attention is constrained by a local context window and gradually forgets earlier entities and attributes;
(ii) \textit{interactive prompt adaptation}~\cite{streamingt2v,longlive}, where users expect to guide evolving generation through streaming inputs;
(iii) \textit{inefficient context scaling}~\cite{deepforcing,lovic}, where simply enlarging the local window neither preserves critical memory nor exploits the computational advantage of AR models over bidirectional ones.
These failures reveal the same problem: the model lacks an explicit state variable for persistent entities, so visual evidence stored as positions, frames, or generic features cannot reliably bind a character to its evolving attributes across prompt switches~\cite{longvideosurvey}.

As illustrated in Fig.~\ref{fig:fig1_teaser}, existing methods can be viewed through two representative memory paradigms.
\textbf{\textit{LongLive}}~\cite{longlive}, inspired by StreamingLLM~\cite{streamingllm}, mitigates quality degradation in streaming rollouts by introducing sink tokens: it always preserves the first chunk as persistent anchors and adopts streaming training to stabilize autoregressive generation. This design improves long-term consistency, but memory fixed to the first chunk discards information accumulated during generation, which can cause memory loss and visual degradation. This reveals the first challenge:
long memory should dynamically preserve critical information rather than rely on a fixed global anchor.
\textbf{\textit{MemFlow}}~\cite{memflow} introduces explicit memory frames within a dynamic memory bank, retrieving and updating semantically relevant historical frames according to the current prompt. While this enables more adaptive historical recall, it mainly treats memory as a streaming retrieval component without systematic organization.
Moreover, increasing the memory or context window can introduce additional inference overhead. This reveals the second challenge:
a practical streaming memory framework must maintain a balance between information preservation and efficient AR generation.

Building on these observations, we propose \textit{\textbf{IAMFlow}}, a training-free framework for narrative autoregressive video generation.
To address identity forgetting and memory loss, we organize an identity-aware memory to introduce explicit entity management into the autoregressive generation process, enabling the model to recall persistent entities and attributes across prompt transitions.
To address the efficiency challenge, we further adopt a systematic inference acceleration pipeline that makes the proposed memory framework computationally practical.
Since existing benchmarks do not fully cover streaming multi-prompt narrative scenarios, we also introduce \textit{\textbf{NarraStream-Bench}}, a benchmark designed to address these evaluation gaps.
The framework, acceleration design, benchmark construction, and evaluation protocol are presented in Sec.~\ref{sec:method_id_memory}, Sec.~\ref{sec:method_acceleration}, and Sec.~\ref{sec:method_narrstream}.

In summary, our contributions are as follows:
\begin{itemize}
  \item We introduce \textbf{IAMFlow}, a training-free identity-aware memory framework that explicitly organizes historical information around persistent entities and attributes, enabling reliable identity preservation across evolving prompt transitions.
  \item We design a systematic inference acceleration pipeline to make the framework computationally practical, combining asynchronous visual verification, adaptive prompt transition, and model quantization to preserve long-term consistency without sacrificing generation speed.
  \item We introduce \textbf{NarraStream-Bench}, a modern benchmark suite for assessing long-term consistency in narrative streaming video generation. Extensive experiments and ablation studies demonstrate that IAMFlow achieves superior performance across various metrics while enabling more efficient inference.
\end{itemize}

\section{Related Work}

\textbf{Autoregressive Video Generation}
Bidirectional video diffusion models~\cite{wan21,cogvideox} are constrained to one-shot, non-interactive short video generation, whereas autoregressive (AR) video models predict future frames conditioned on previous context.
Early methods~\cite{pyramidalflow,nova} adopt Teacher Forcing (TF) for training, using ground-truth historical frames as conditioning input, which suffers from a severe train-test gap during long-horizon rollout. Subsequent systems~\cite{skyreelsv2,magi1} adopt Diffusion Forcing (DF)~\cite{diffusionforcing}, which injects noise into the conditioning context during training to mitigate error accumulation. Building on this foundation, recent works further explore causal modeling and efficient generation. CausVid~\cite{causvid} distills a pretrained bidirectional model into a few-step causal generator.
Self Forcing~\cite{selfforcing} simulates autoregressive rollouts during training, reducing the train-test discrepancy and error propagation.
Follow-up forcing-style methods~\cite{rollingforcing,selfforcingpp,rewardforcing,contextforcing} improve robustness to long-horizon rollout.
In addition, PA-VDM~\cite{pavdm} introduces progressive frame-wise noise scheduling, while StreamingT2V~\cite{streamingt2v} introduces short-term and long-term modules for extendable generation.

\textbf{Memory Mechanisms in Video Generation}
A central challenge in autoregressive long video generation is preserving relevant history over long horizons under bounded computation and memory.
Existing methods address this problem through three main memory designs. Compressed-context approaches reduce the growing video history into a compact conditioning signal: FramePack~\cite{framepack} packs historical frames into a fixed context budget, FAR~\cite{far} combines fine recent context with coarser distant history, and LoViC~\cite{lovic} and PFP~\cite{pfp} compress long histories into latent or retrievable memory representations. Parameterized memory methods instead store history in learnable internal states. TTT-Video~\cite{ttt-video} updates sequence-specific fast weights at test time, SlowFast-VGen~\cite{slowfastvgen} stores episodic information in inference-time temporal LoRA parameters, and VideoSSM~\cite{videossm} combines local autoregressive context with state-space memory.
Retrieval-based methods select relevant history during generation. MemoryPack-based generation~\cite{pfvg} retrieves and fuses text-visual history, while InfLVG~\cite{inflvg} learns inference-time context selection under a fixed memory budget. LongLive~\cite{longlive} maintains a streaming cache with frame-level attention sinks and KV-recache, and MemFlow~\cite{memflow} retrieves semantically relevant memory frames from a dynamic memory bank.

\textbf{Benchmarks for Video Generation}
Existing video-generation benchmarks~\cite{vbench,vbenchpp,blockvid}, along with recent narrative long-video evaluations~\cite{moviebench,narrlv}, cover many aspects of long-horizon video generation, but they do not fully cover streaming multi-prompt narrative scenarios.
VBench and VBench++ mainly provide broad quality and semantic metrics for general video generation, while VBench-Long evaluates temporal consistency within a single long-form generation by splitting videos into semantically consistent clips and using a slow-fast protocol.
BlockVid focuses on minute-long video generation, and MovieBench and NarrLV move evaluation toward movie-level and narrative settings.
These benchmarks do not explicitly test continual adaptation to evolving prompts, instruction following under state changes between segments, or long-range entity recall after reappearance.

\section{Method}\label{sec:method}
We present IAMFlow, a training-free identity-aware memory framework that builds an explicit ID-centric memory bank to improve temporal consistency in long video generation (Sec.~\ref{sec:method_id_memory}).
To make the proposed memory framework computationally practical, we further develop a systematic inference acceleration pipeline for efficient streaming generation (Sec.~\ref{sec:method_acceleration}).
To evaluate the proposed framework, we introduce NarraStream-Bench, a benchmark suite for assessing long-term consistency in narrative streaming video generation (Sec.~\ref{sec:method_narrstream}).
\begin{figure}[t]
  \centering
  \includegraphics[width=\linewidth]{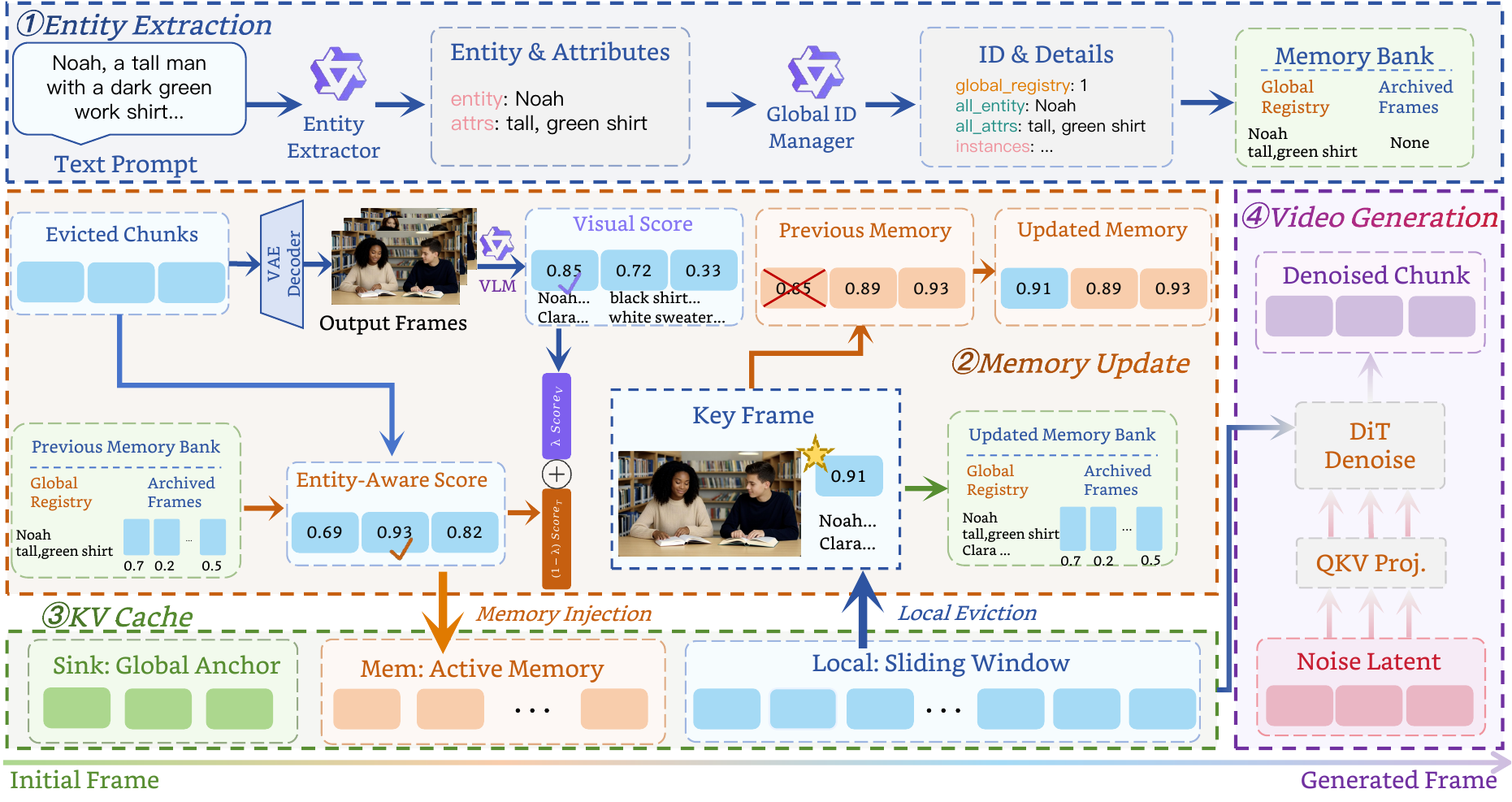}
  \caption{\textbf{Framework of IAMFlow}.
  (1) An LLM extracts entities and attributes from each prompt and assigns persistent global IDs to build an ID-centric memory bank.
  (2) At each chunk generation, the memory bank retrieves frames associated with critical entities.
  (3) During autoregressive generation, evicted local chunks are decoded and scored, while asynchronous VLM verification updates attributes and selects reliable keyframes.
  (4) The latent noise is concatenated with the organized KV cache and fed into the DiT to generate the denoised chunk.}
  \label{fig:fig2_method}
\end{figure}

\textbf{Formulation of Memory-aware Autoregressive Video Generation.}
Given a stream of prompts $\{p_t\}_{t=1}^{T}$ and corresponding generated video units $\{x_t\}_{t=1}^{T}$, where each $x_t$ can denote a frame or latent chunk, memory-aware autoregressive video diffusion models factorize generation as
\begin{equation}\label{eq:memory_aware_ar}
 p_\theta(x_{1:T}\mid p_{1:T}) = \prod_{t=1}^{T} p_\theta\!\left(x_t \mid \mathcal{L}_t,\,\mathcal{M}_t,\,p_t\right).
\end{equation}
Here $p_t$ is the active prompt for generation unit $x_t$, $\mathcal{L}_t$ is the recent local context, instantiated as $x_{t-W_{\mathrm{loc}}+1:t-1}$ with local window size $W_{\mathrm{loc}}$, and $\mathcal{M}_t$ is the compact long-term memory selected from a method-specific memory state $\mathcal{S}_t$ under a budget $B$.
Under this formulation, existing methods can be organized into a \textbf{Memory Organization Paradigm}, where they share the same autoregressive objective in Eq.~\ref{eq:memory_aware_ar} but differ fundamentally in how they represent and update $\mathcal{M}_t$:

\textbf{Positional Anchoring.}
Early approaches such as LongLive~\cite{longlive} instantiate $\mathcal{M}_t$ with persistent sink tokens anchored solely to the initial temporal positions:
\begin{equation}
 p_{\mathrm{LL}}(x_{1:T}\mid p_{1:T})
 = \prod_{t=1}^{T} p_\theta\!\left(x_t \mid x_{t-W_{\mathrm{loc}}+1:t-1},\,\mathrm{Sink}(x_{1:s}),\,p_t\right).
\end{equation}
While mitigating degradation in early generation steps, this static strategy fails to capture the dynamic evolution of narrative events beyond the first chunk.

\textbf{Token-Level Retrieval.}
To enable dynamic updates, MemFlow~\cite{memflow} uses a dynamic memory bank, organizing $\mathcal{M}_t$ based on token-level relevance retrieved via the current prompt:
\begin{equation}
 p_{\mathrm{MF}}(x_{1:T}\mid p_{1:T})
 = \prod_{t=1}^{T} p_\theta\!\left(x_t \mid x_{t-W_{\mathrm{loc}}+1:t-1},\,\mathrm{Retrieve}(\mathcal{F}_t,p_t),\,p_t\right).
\end{equation}
Although this design supports adaptive memory updates through $\mathcal{F}_t$, it still operates on frame-level features rather than explicit entity identities. This makes it prone to \emph{feature conflation}: different entities with similar visual or semantic attributes may be confused, while less prominent subjects may not be reliably retrieved when targeted recall is needed.

\subsection{Identity-Aware Memory Preservation}
\label{sec:method_id_memory}
To address the limitations above, IAMFlow elevates memory organization from low-level tokens to high-level symbolic entities.
Specifically, we instantiate $\mathcal{M}_t$ as an ID-centric memory state, which enables targeted recall of persistent entities and reduces identity conflation in previous memory designs.

We propose an \textbf{Identity-Aware Memory} mechanism (Fig.~\ref{fig:fig2_method}) that maintains a memory bank, including a global entity registry $\mathcal{R}$, a frame archive $\mathcal{F}$, and an active memory buffer $m^{\mathrm{id}}$.
At each prompt $p_t$, the mechanism executes four phases:
\textbf{(i)}~an LLM~\cite{qwen3} extracts entities and assigns persistent global IDs;
\textbf{(ii)}~the memory bank retrieves frames covering all active identities;
\textbf{(iii)}~during generation, evicted chunks are scored and the best frame is archived;
\textbf{(iv)}~a VLM~\cite{qwen3vl} asynchronously scores visual quality and corrects attribute drift.
IAMFlow therefore instantiates Eq.~\ref{eq:memory_aware_ar} as:
\begin{equation}\label{eq:iamflow_memory}
 p_{\mathrm{IAM}}(x_{1:T}\mid p_{1:T})
 = \prod_{t=1}^{T} p_\theta\!\left(x_t \mid x_{t-W_{\mathrm{loc}}+1:t-1},\,\mathrm{IDRetrieve}(\mathcal{F}_t,\mathcal{E}_t,\mathcal{R}_t),\,p_t\right).
\end{equation}
Here, $\mathcal{R}_t$ and $\mathcal{F}_t$ together define the IAMFlow identity-aware memory state, while $\mathcal{E}_t$ denotes the entity descriptors parsed from the current prompt and aligned with the global registry before identity-aware retrieval.

\textbf{Entity Extraction and ID Assignment.}
At the first chunk of each prompt $p_t$, an LLM parses the text into structured entity descriptors $\mathcal{E}_t = \{(e_j, \mathbf{a}_j)\}$, where $e_j$ is an entity name and $\mathbf{a}_j$ lists its stable visual attributes.
Each entity is linked to a persistent ID in the global registry $\mathcal{R} = \{g_k: (\mathrm{name}_k, \mathbf{A}_k)\}$ via a matching procedure for its novelty: first, explicit novelty markers trigger immediate allocation of a new ID. Otherwise, the LLM compares the current attribute set $\mathbf{a}_j$ with the registered attribute sets $\{\mathbf{A}_k\}$ to resolve ambiguous references. If a confident match is found, the entity is linked to the corresponding existing ID; if not, a new ID is allocated.
The registry is then updated with the new observation and stored in the memory bank.

\textbf{Identity-aware Frame Scoring, Archival, and Retrieval.}
The memory bank maintains a growing frame archive $\mathcal{F}$. During chunk-wise generation, when the oldest chunk is evicted from the local window, we score its frames and archive the best one.
To prioritize content tied to active identities, we construct an entity-token weight
vector $\boldsymbol{\omega}\in\mathbb{R}^{S}$ over the prompt tokens according to the entity names
and attributes maintained in the memory bank, and normalize it as
$\tilde{\omega}_u=\omega_u/\sum_{v=1}^{S}\omega_v$.
For layer $l \in \mathcal{L}$ and attention head $h \in \mathcal{H}$,
we aggregate the cached text keys and the visual keys for frame $f$ as
$\bar{\mathbf{r}}^{l,h}_{\mathrm{id}}=\sum_{u=1}^{S}\tilde{\omega}_u
\mathbf{K}^{l,\mathrm{text}}_{u,h}$ and
$\bar{\mathbf{k}}^{l,h}_{f}=\frac{1}{n_f}\sum_{v=1}^{n_f}
\mathbf{K}^{l,\mathrm{vis}}_{f,v,h}$. The entity score is computed by
\begin{equation}\label{eq:entity_score}
s_{\mathrm{entity}}(f)
=
\sum_{l\in\mathcal{L}}\beta_l
\frac{1}{H}
\sum_{h=1}^{H}
\frac{
\langle
\bar{\mathbf{r}}^{l,h}_{\mathrm{id}},
\bar{\mathbf{k}}^{l,h}_{f}
\rangle
}{\sqrt{d}}.
\end{equation}
This score reuses the generator's internal compatibility between text-conditioned keys and visual keys as a lightweight saliency proxy for ranking memory frames that best preserve identity.
The entity score is then fused with the VLM-based score as $s(f) = (1{-}\lambda)\,s_{\mathrm{entity}}(f) + \lambda\,s_{\mathrm{visual}}(f)$.
The best frame is stored in $\mathcal{F}$ along with its entity annotations and KV cache.
We formulate the ID retrieval as a maximum coverage problem under a memory budget. Exploiting the inherent submodularity of this NP-hard problem, we adopt a greedy approximation. Each archived frame $f$ is associated with an entity set $\mathcal{G}(f)$ and a score, and we iteratively select the frame with the highest score until all current IDs $\{g_j\}$ are covered. The KV caches of the selected frames are then assembled into the active memory $m^{\mathrm{id}}_t$ and injected into the model.

\textbf{Asynchronous Visual Verification and Correction.}
Following each chunk, the denoised latent is asynchronously decoded into pixels via the VAE.
Then a VLM scores each decoded frame for visual fidelity to the target entities and returns $s_{\mathrm{visual}}(f) \in [0,1]$ for the fusion score above. For the first chunk of each prompt, the VLM also verifies entity attributes from the rendered pixels and writes corrected observations back to the global registry,
thereby reducing error accumulation across prompt switches.
Detailed runtime implementation and deterministic prompt templates are provided in Appendices~\ref{app:memory_implement} and~\ref{app:agent_prompts}.

\subsection{Systematic Inference Acceleration}\label{sec:method_acceleration}
Prior methods in autoregressive long video generation are bottlenecked by two recurring costs: full KV cache recomputation at every prompt switch and synchronous frame decoding~\cite{quantvideogen}.
We introduce three complementary strategies that eliminate both bottlenecks, enabling IAMFlow to achieve faster end-to-end inference than prior methods even with the additional components.

\textbf{Adaptive Prompt Transition.}
In prior autoregressive generation methods~\cite{longlive,memflow}, each prompt switch requires full recaching: all context frames must be re-encoded under the new text condition before generation can resume.
We introduce an \textbf{Adaptive Prompt Transition} (APT) strategy that eliminates this cost by smoothing the prompt switch inside the cross-attention key-value conditioning space.
We view APT as smoothing the text condition across segment boundaries.
Instead of switching abruptly from one prompt to the next, APT interpolates the cross-attention keys and values used by the DiT, so the denoiser receives a gradual change in conditioning.
This leads to smoother denoising directions across neighboring chunks.
At the switch point, the current cross-attention keys and values $(\mathbf{K}_{\mathrm{old}}, \mathbf{V}_{\mathrm{old}})$ are cached, and subsequent chunks use blended states:
\begin{equation}\label{eq:apt_blend}
  \mathbf{K} = (1{-}\alpha)\,\mathbf{K}_{\mathrm{old}} + \alpha\,\mathbf{K}_{\mathrm{new}}, \quad
  \mathbf{V} = (1{-}\alpha)\,\mathbf{V}_{\mathrm{old}} + \alpha\,\mathbf{V}_{\mathrm{new}},
\end{equation}
where $\alpha$ increases from 0 to 1 over a transition window following a cosine schedule $\alpha(\tau) = \tfrac{1}{2}(1-\cos\pi \tau)$.
To adapt the transition length to the magnitude of the prompt change, we compute the cosine distance $d$ between the averaged prompt embeddings of two adjacent segments, reusing the existing outputs of the text encoder at no additional cost, and set the transition window length to $W_{\mathrm{apt}} = W_{\min} + d\cdot(W_{\max} - W_{\min})$, rounded to the nearest chunk boundary.
As a result, minor prompt edits use shorter transitions, while larger narrative changes trigger longer blending windows.
By introducing prompt changes gradually over a short temporal window, APT reduces boundary flicker, identity drift, and transition artifacts.
It also avoids the full recache pass required by hard prompt switching, replacing each switch with a lightweight blend of the corresponding key and value.

\textbf{Asynchronous Verification Pipeline.}
Rather than interrupting autoregressive generation, we integrate the VAE decoding and VLM scoring loop with the local window eviction mechanism. Once a mature chunk is evicted, a background thread decodes it and runs VLM verification in parallel with DiT denoising for subsequent chunks. This keeps only visually reliable, identity-consistent frames in long-term memory while overlapping verification with generation.

\textbf{Efficient Deployment.}
To further accelerate autoregressive generation with identity-aware memory, we deploy the DiT, VAE, and text encoder on one GPU, while serving the LLM and VLM on another GPU via vLLM~\cite{vllm}. We further replace the default Wan VAE~\cite{wan21} with a Wan2.1-adapted TinyVAE based on Tiny AutoEncoder~\cite{taehv}, and quantize all DiT linear layers to FP8~\cite{fp8formats} following LightX2V~\cite{lightx2v}. These optimizations provide additional speedup.
\subsection{NarraStream-Bench: Evaluating Narrative Streaming Video Generation}\label{sec:method_narrstream}
\begin{figure}[t]
  \centering
  \includegraphics[width=\linewidth]{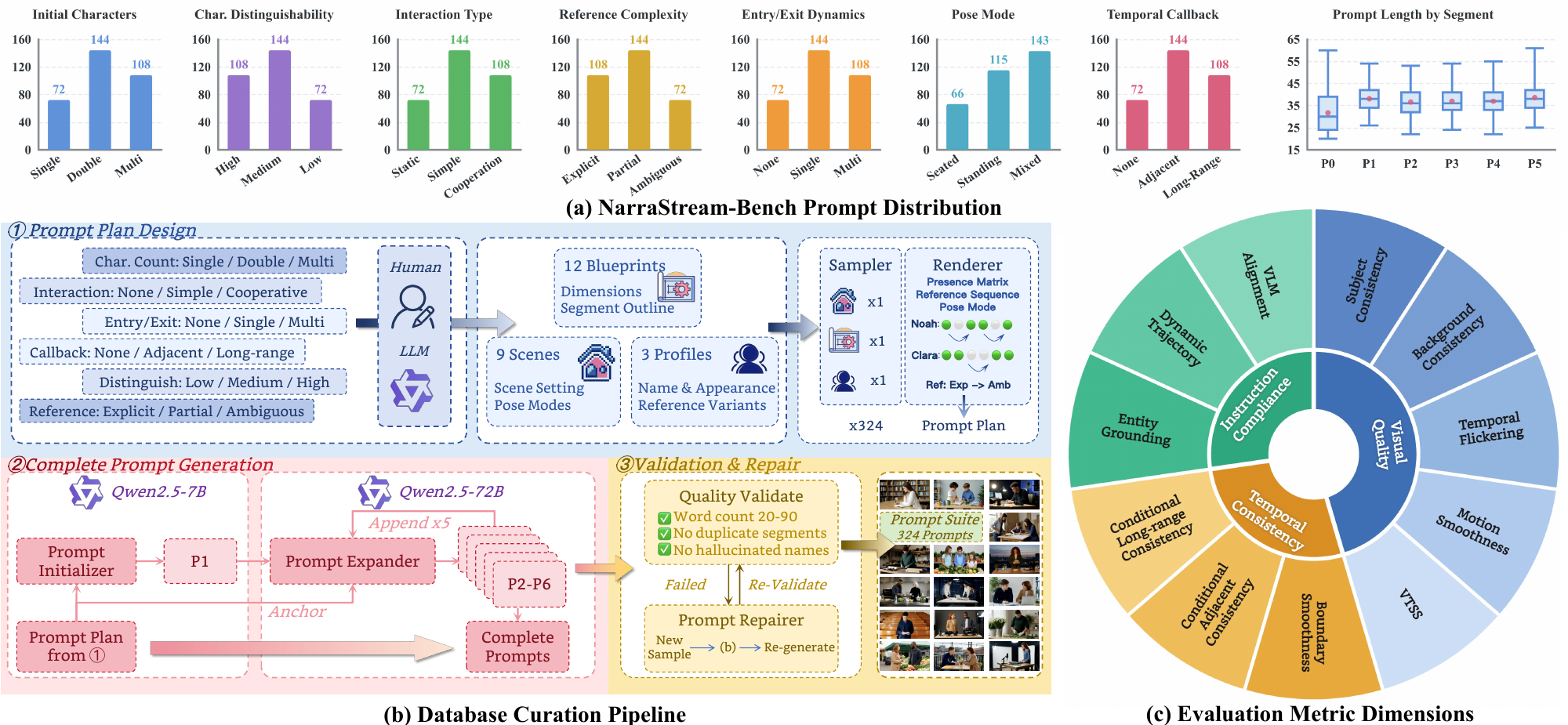}
  \captionsetup{justification=centering}
  \caption{\textbf{Overview of NarraStream-Bench}.}
  \label{fig:fig3_benchmark}
\end{figure}

To address the lack of benchmarks for continual prompt adaptation, instruction compliance between segments, and long-term consistency in streaming narrative video generation, we introduce \textbf{NarraStream-Bench}, a structured benchmark comprising 324 narrative scripts, each spanning 60 seconds and divided into six consecutive segments, together with metrics specifically designed for streaming narrative video generation (see Fig.~\ref{fig:fig3_benchmark}).
Table~\ref{tab:tb1_benchmark_comparison} compares the evaluation protocols of NarraStream-Bench and related representative benchmarks.
We provide the implementation details for database construction and evaluation metrics in Appendices~\ref{app:narrstream_construction} and~\ref{app:narrstream_metrics}, respectively.

\textbf{Test Suite Construction.}
As shown in Fig.~\ref{fig:fig3_benchmark}(b), we define six challenge dimensions for narrative streaming video generation and build a prompt database of scene templates, narrative blueprints, and character profiles through human drafting and LLM refinement. We then sample prompt plans from this database, augment each plan with a presence matrix and reference sequence between segments, and generate six coherent segments following the formats of Wan2.1 and LongLive. After quality validation and repair for logical or formatting issues, the final suite contains 324 prompt sequences.

\begin{table}[t]
  \centering
  \caption{\textbf{Comparison of related long-video generation benchmarks}. Note that the custom VBench-Long protocol only supports metrics for visual quality.}
  \label{tab:tb1_benchmark_comparison}
  \scriptsize
  \setlength{\tabcolsep}{3pt}
  \renewcommand{\arraystretch}{1.00}
  \begin{tabularx}{\linewidth}{@{}
    >{\hsize=1.35\hsize\linewidth=\hsize\raggedright\arraybackslash}X
    >{\hsize=0.65\hsize\linewidth=\hsize\centering\arraybackslash}X
    >{\hsize=0.65\hsize\linewidth=\hsize\centering\arraybackslash}X
    >{\hsize=0.65\hsize\linewidth=\hsize\centering\arraybackslash}X
    >{\hsize=1.45\hsize\linewidth=\hsize\centering\arraybackslash}X
    >{\hsize=1.45\hsize\linewidth=\hsize\raggedright\arraybackslash}X
    >{\hsize=0.80\hsize\linewidth=\hsize\centering\arraybackslash}X
  @{}}
    \toprule
    Benchmark & VQ & TC & IC & \mbox{Prompt Type} & Aggregation Strategy & Year \\
    \midrule
    VBench-Long~\cite{vbenchpp} & $\checkmark$ & $\times$ & $\times$ & Single & Slow-Fast Avg. & 2024 \\
    LV-Bench~\cite{blockvid} & $\checkmark$ & $\checkmark$ & $\times$ & Single & VDE & 2025 \\
    NarrLV~\cite{narrlv} & $\times$ & $\checkmark$ & $\checkmark$ & Single & TNA-based QA & 2025 \\
    \midrule
    \rowcolor{blue!8}
    \textbf{NarraStream-Bench} & $\checkmark$ & $\checkmark$ & $\checkmark$ & Multi & Narrative-Aware & 2026 \\
    \bottomrule
  \end{tabularx}
\end{table}

\textbf{Evaluation Metrics.}
We define 11 metrics across three complementary groups, as illustrated in Fig.~\ref{fig:fig3_benchmark}(c):
1) \textit{\textbf{Visual Quality}} measures visual fidelity with five metrics: Subject Consistency (SC), Background Consistency (BC), Temporal Flickering (TF), Motion Smoothness (MS), and VTSS;
2) \textit{\textbf{Temporal Consistency}} measures transition smoothness and long-term consistency with three metrics: Boundary Smoothness (BS), Conditional Adjacent Consistency (CAC), and Conditional Long-range Consistency (CLC);
3) \textit{\textbf{Instruction Compliance}} measures instruction following with three metrics: Entity Grounding (EG), Dynamic Trajectory (DT), and VLM Alignment (VLM)~\cite{videoscore,t2vcompbench}.
Rather than averaging scores uniformly, we use an LLM to assign segment weights for each metric based on the prompt sequence.
The final score for each group is computed as an importance-weighted average across segments, and the overall score is the arithmetic mean of the three group scores.

\section{Experiments}
\label{sec:experiments}
\begin{figure}[t]
  \centering
\includegraphics[width=\linewidth]{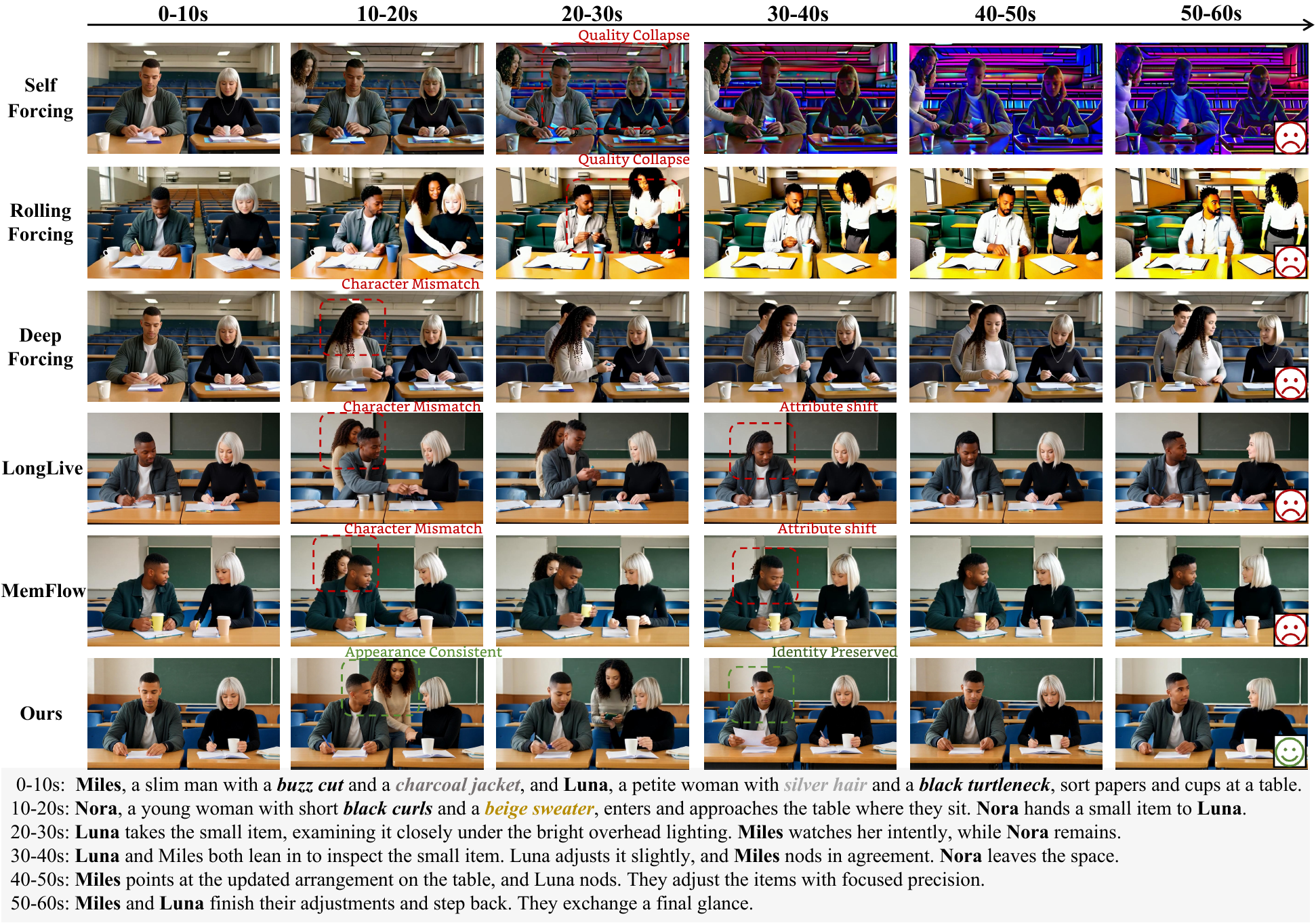}
  \caption{\textbf{Qualitative results of IAMFlow.} Our method keeps identities and attributes consistent across the 60-second video, while other baselines show varying degrees of identity shift and attribute drift.}

  \label{fig:fig4_main_comparison}
  \vspace{-8pt}
\end{figure}

\textbf{Implementation Details.}
IAMFlow uses the MemFlow~\cite{memflow} weights to generate 60-second videos from six successive prompts at $832{\times}480$ resolution, adding the proposed identity-aware memory and inference pipeline without additional training. The KV cache is organized as $N_s\!=\!3$ sink latent frames, an active memory buffer of $N_m\!\in\![2,4]$ retrieved latent frames, and a sliding local window of $N_l\!=\!9$ latent frames; we deploy Qwen3-4B-Instruct~\cite{qwen3} and Qwen3-VL-2B-Instruct~\cite{qwen3vl} as the LLM and VLM via vLLM~\cite{vllm}. Keyframe scoring uses $\lambda\!=\!0.3$, Adaptive Prompt Transition uses $W_{\min}\!=\!3$ and $W_{\max}\!=\!15$ chunks, and all experiments are conducted on NVIDIA H20 GPUs.

\textbf{Baselines.}
We benchmark IAMFlow against comparable open-source video generation models, including Self Forcing~\cite{selfforcing}, Rolling Forcing~\cite{rollingforcing}, Deep Forcing~\cite{deepforcing}, LongLive~\cite{longlive}, and MemFlow~\cite{memflow}. We evaluate all methods on \textbf{NarraStream-Bench}, report 11 metrics scaled by 100, and additionally measure human alignment and inference efficiency under the same evaluation budget.

\subsection{Comparison on NarraStream-Bench}
As several representative baselines were originally designed for single-prompt rollouts, we extend Self Forcing, Rolling Forcing, and Deep Forcing to the streaming video generation setting; Appendix~\ref{app:baseline_implementation} provides the details. We then compare our method against all baselines.
Qualitative and quantitative results are shown in Fig.~\ref{fig:fig4_main_comparison} and Table~\ref{tab:tb2_narrstream_main}.
IAMFlow achieves the best overall score, with the most pronounced gains on the metrics that stress entity-state memory: Boundary Smoothness, Conditional Adjacent Consistency, and Conditional Long-range Consistency.
In contrast, LongLive and MemFlow show limited long-term consistency, supporting our central claim that narrative long video generation benefits from remembering persistent entities.
The adapted forcing baselines lag further behind, as they mainly preserve recent context and lack a memory mechanism for long video generation.
Further qualitative comparisons and discussions of challenging failure cases and limitations are provided in Appendices~\ref{app:more_qual_details} and~\ref{app:limitations}.

\begin{table}[t]
\centering
\caption{\textbf{Quantitative comparison for the multi-prompt 60-second setting on NarraStream-Bench}. Self Forcing, Rolling Forcing, and Deep Forcing are adapted for the experiment. The best results are highlighted in bold, and the second-best results are underlined.}
\label{tab:tb2_narrstream_main}
\small
\setlength{\tabcolsep}{3.5pt}
\renewcommand{\arraystretch}{1.08}
\resizebox{\linewidth}{!}{%
\begin{tabular}{lcccccccccccc}
\toprule
\multirow{2}{*}{Model} & \multicolumn{5}{c}{Visual Quality $\uparrow$} & \multicolumn{3}{c}{Temporal Consistency $\uparrow$} & \multicolumn{3}{c}{Instruction Compliance $\uparrow$} & \multirow{2}{*}{Overall $\uparrow$} \\
\cmidrule(lr){2-6}\cmidrule(lr){7-9}\cmidrule(lr){10-12}
 & SC & BC & TF & MS & VTSS & BS & CAC & CLC & EG & DT & VLM \\
\midrule
Self Forcing~\cite{selfforcing} & 80.85 & 82.05 & 86.29 & 56.17 & 30.39 & 60.25 & 50.57 & 39.54 & 55.63 & 51.98 & 50.96 & 56.71 \\
Rolling Forcing~\cite{rollingforcing} & 88.16 & 85.93 & 80.55 & 41.41 & 52.59 & \underline{61.19} & 68.83 & 62.18 & 58.45 & 56.37 & 52.03 & 63.14 \\
Deep Forcing~\cite{deepforcing} & 92.27 & 90.74 & 88.54 & 50.61 & 73.99 & 48.10 & 80.09 & 77.28 & 60.71 & 62.87 & 59.35 & 69.57 \\
LongLive~\cite{longlive} & 95.36 & \underline{93.32} & \underline{90.27} & \underline{57.76} & 77.29 & 49.19 & 83.06 & \underline{89.97} & \textbf{62.40} & 63.33 & \underline{62.22} & \underline{73.17} \\
MemFlow~\cite{memflow} & \underline{95.37} & 93.21 & 90.23 & 57.37 & \textbf{77.67} & 47.06 & \underline{83.92} & 89.71 & 61.93 & \underline{64.14} & 60.85 & 72.88 \\
\midrule
\rowcolor{blue!8}
\textbf{IAMFlow} & \textbf{95.88} & \textbf{93.43} & \textbf{90.36} & \textbf{59.49} & \underline{77.56} & \textbf{65.03} & \textbf{86.07} & \textbf{91.25} & \underline{62.03} & \textbf{64.56} & \textbf{62.61} & \textbf{75.73} \\
\bottomrule
\end{tabular}%
}
\end{table}

\subsection{Evaluation on VBench-Long}
\begin{wraptable}{r}{0.52\linewidth}
\vspace{-0.8em}
\centering
\caption{\textbf{Evaluation metrics on VBench-Long}. All scores are multiplied by 100. Best results are highlighted in bold, and second-best results are underlined.}
\label{tab:tb3_vbench_long_metrics}
\scriptsize
\setlength{\tabcolsep}{3.1pt}
\setlength{\extrarowheight}{0.5pt}
\renewcommand{\arraystretch}{1.08}
\setlength{\aboverulesep}{0.35ex}
\setlength{\belowrulesep}{0.45ex}
\begin{tabularx}{\linewidth}{@{}l
  >{\centering\arraybackslash}X
  >{\centering\arraybackslash}X
  >{\centering\arraybackslash}X@{}}
\toprule
Metric & LongLive & MemFlow & \textbf{IAMFlow} \\
\midrule
Subject Cons. & \underline{98.956} & 98.946 & \textbf{99.006} \\
Background Cons. & \underline{96.394} & 96.170 & \textbf{96.554} \\
Temporal Flick. & \underline{99.402} & 99.386 & \textbf{99.438} \\
Motion Smooth. & \textbf{17.202} & 16.440 & \underline{16.522} \\
Overall Cons. & \underline{59.420} & 57.667 & \textbf{59.630} \\
Dynamic Degree & \underline{75.895} & 75.726 & \textbf{75.942} \\
\bottomrule
\end{tabularx}
\vspace{-1.2em}
\end{wraptable}

We also evaluate IAMFlow on VBench-Long~\cite{vbenchpp}, a benchmark for long video generation.
Since VBench-Long is designed mainly for single-prompt evaluation, we follow LongLive~\cite{longlive} and MemFlow~\cite{memflow} and use the custom-prompt protocol, with prompts drawn from the NarraStream-Bench database.
Under this protocol, VBench-Long supports six metrics: Subject Consistency, Background Consistency, Temporal Flickering, Motion Smoothness, Overall Consistency, and Dynamic Degree.
As shown in Table~\ref{tab:tb3_vbench_long_metrics}, IAMFlow achieves the best performance on five of the six metrics and remains competitive on Motion Smoothness.

\subsection{Ablation Studies}
\begin{figure}[t]
  \centering
\includegraphics[width=\linewidth]{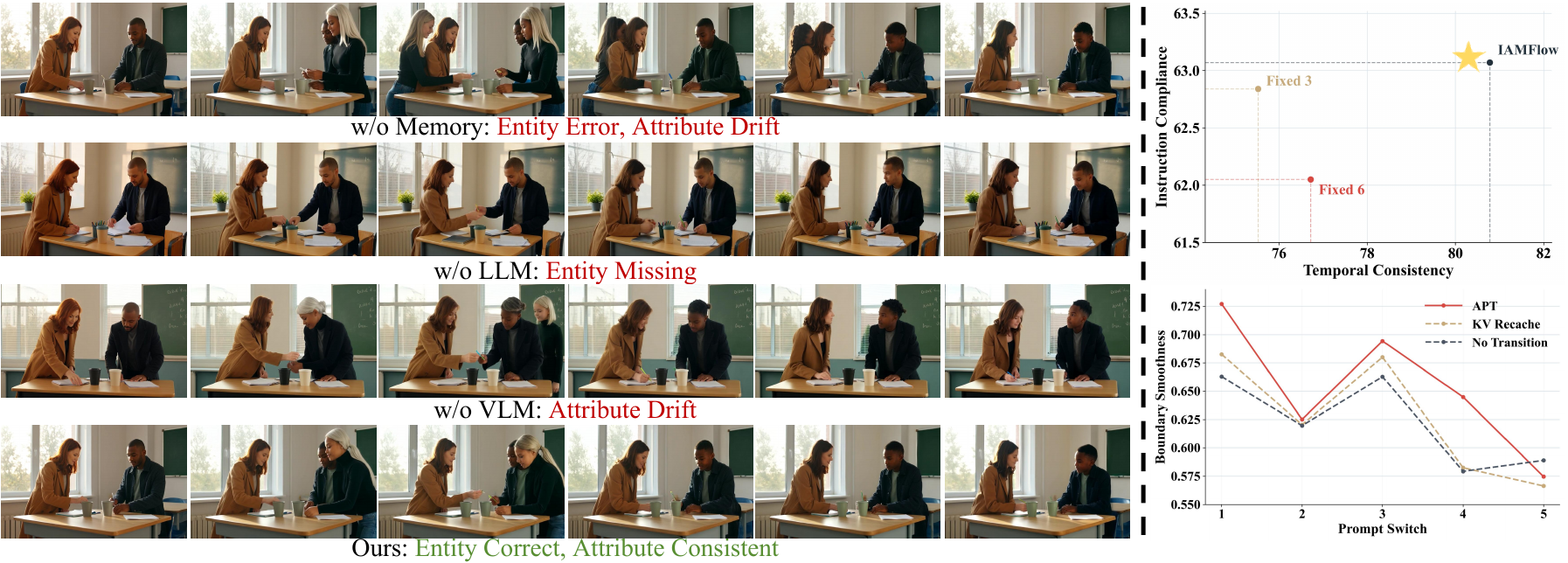}
  \caption{\textbf{Left}: Qualitative results of ablation experiments. \textbf{Right}: \textit{Top}, the dynamic strategy achieves the best balance between temporal consistency and instruction compliance; \textit{Bottom}, Adaptive Prompt Transition (APT) enables smoother boundary transitions.}
  \label{fig:fig5_ablation}
  \vspace{-8pt}
\end{figure}

\begin{table}[t]
\centering
\caption{\textbf{Unified ablation studies of IAMFlow}. The best
results are highlighted in bold, and the second-best results are underlined.}
\label{tab:tb4_ablation_combined}
\small
\setlength{\tabcolsep}{3pt}
\renewcommand{\arraystretch}{1.05}
\setlength{\aboverulesep}{0.25ex}
\setlength{\belowrulesep}{0.35ex}
\begin{tabularx}{\linewidth}{@{}l
  >{\hsize=.55\hsize\centering\arraybackslash}X
  >{\hsize=.90\hsize\centering\arraybackslash}X
  >{\hsize=1.25\hsize\centering\arraybackslash}X
  >{\hsize=1.30\hsize\centering\arraybackslash}X@{}}
\toprule
Model & Overall $\uparrow$ & Visual Quality $\uparrow$ & Temporal Consistency $\uparrow$ & Instruction Compliance $\uparrow$ \\
\midrule
\rowcolor{gray!12}
\multicolumn{5}{l}{\textit{Memory Mechanism}} \\
w/o VLM verification & \underline{75.41} & 83.17 & \underline{80.22} & 62.83 \\
w/o LLM planning & 75.26 & 83.09 & 80.12 & 62.58 \\
w/o ID memory & 74.28 & 82.74 & 77.85 & 62.25 \\
\midrule
\rowcolor{gray!12}
\multicolumn{5}{l}{\textit{Memory Retrieval}} \\
Random & 73.10 & 82.39 & 73.89 & \underline{63.02} \\
Visual-Only & 71.35 & 80.94 & 71.08 & 62.04 \\
Semantic-Only & 73.49 & 82.70 & 74.85 & 62.91 \\
\midrule
\rowcolor{gray!12}
\multicolumn{5}{l}{\textit{Memory Allocation}} \\
Fixed 3 & 73.69 & 82.71 & 75.53 & 62.84 \\
Fixed 6 & 74.00 & \underline{83.24} & 76.72 & 62.05 \\
\midrule
\rowcolor{gray!12}
\multicolumn{5}{l}{\textit{Prompt Transition}} \\
Full KV recache & 75.27 & 83.12 & 79.71 & 62.97 \\
No transition & 75.10 & 83.01 & 79.42 & 62.88 \\
\midrule
\rowcolor{blue!8}
\textbf{Ours} & \textbf{75.73} & \textbf{83.34} & \textbf{80.78} & \textbf{63.07} \\
\bottomrule
\end{tabularx}
\end{table}

We evaluate four components in IAMFlow: \textbf{Memory Mechanism}, \textbf{Memory Retrieval}, \textbf{Memory Allocation}, and \textbf{Prompt Transition}. Table~\ref{tab:tb4_ablation_combined} and Fig.~\ref{fig:fig5_ablation} show that the full IAMFlow achieves the best overall performance. Removing the identity-aware memory causes the largest degradation, confirming that explicit long-term memory is critical to identity preservation and temporal consistency. LLM planning and VLM verification provide complementary gains through structured entity decomposition and visual grounding. Identity-aware retrieval outperforms random, visual-only, and semantic-only retrieval, while dynamic allocation and Adaptive Prompt Transition further improve the balance between identity coverage, instruction compliance, and transition smoothness.

\subsection{Efficiency Comparison}
We incorporate a systematic inference acceleration pipeline into IAMFlow, enabling faster end-to-end inference than prior methods despite the additional memory components. As shown in Table~\ref{tab:tb5_efficiency}, results on H20 GPUs show that our method achieves a $1.69\times$ end-to-end speedup compared to Self Forcing in the same multi-prompt setting. This comparison reflects the overall efficiency of our method. Appendix~\ref{app:inference_acceleration} further decomposes the contributions of the pipeline components. Notably, even with the additional memory module, IAMFlow still delivers slightly higher FPS than LongLive.

\begin{table}[t]
\centering
\caption{\textbf{End-to-end efficiency comparison under NVIDIA H20 GPUs}. IAMFlow achieves the highest speedup in the 60-second multi-prompt setting across representative baselines.}
\label{tab:tb5_efficiency}
\small
\setlength{\tabcolsep}{6pt}
\renewcommand{\arraystretch}{1.05}
\setlength{\aboverulesep}{0.25ex}
\setlength{\belowrulesep}{0.35ex}
\begin{tabularx}{\linewidth}{l>{\centering\arraybackslash}X>{\centering\arraybackslash}X>{\centering\arraybackslash}X>{\centering\arraybackslash}X}
\toprule
Method & Training-free & E2E Latency (s) $\downarrow$ & Speedup $\uparrow$ & FPS $\uparrow$ \\
\midrule
Self Forcing & $\times$ & 264.86 & 1.0$\times$ & 4.72 \\
Rolling Forcing & $\times$ & 312.11 & 0.85$\times$ & 4.03 \\
Deep Forcing & $\checkmark$ & 250.62 & 1.06$\times$ & 5.11 \\
LongLive & $\times$ & \underline{218.43} & \underline{1.21$\times$} & \underline{6.14} \\
MemFlow & $\times$ & 241.76 & 1.10$\times$ & 5.36 \\
\midrule
\rowcolor{blue!8}
\textbf{IAMFlow} & $\checkmark$ & \textbf{156.69} & \textbf{1.69$\times$} & \textbf{6.20} \\
\bottomrule
\end{tabularx}
\end{table}
\vspace{-5pt}

\subsection{Evaluator Robustness}
Because IAMFlow uses Qwen3-VL-2B-Instruct for internal visual verification and attribute correction, a natural concern is that the primary NarraStream-Bench judge, Qwen3-VL-30B-A3B-Instruct, may share family-specific visual or textual preferences with the generation-time verifier.
To test whether the evaluation conclusion is sensitive to this same-family evaluator choice, we conduct a cross-judge robustness analysis on LongLive, MemFlow, and IAMFlow.
We keep the generated videos, prompt sequences, sampled frames, metric rubrics, and aggregation protocol unchanged, and replace only the MLLM judge with independent non-Qwen multimodal evaluators.

The analysis focuses on the NarraStream-Bench metrics that directly use large-model judgments: transition selection for CAC, entity-occurrence extraction for CLC, frame-based entity grounding for EG, and VLM Alignment.
As shown in Table~\ref{tab:tb6_benchmark_evaluation_robustness}, IAMFlow remains the best method on CAC, CLC, and VLM Alignment under both non-Qwen judges, and the same trend is preserved under the original Qwen3-VL-30B-A3B judge.
EG is consistently close across judges and does not show a Qwen-specific preference for IAMFlow.
These results indicate that the reported improvements, especially on the VLM Alignment metric, are not an artifact of using a Qwen-family evaluator.

\begin{table}[t]
\centering
\caption{\textbf{Robustness of NarraStream-Bench conclusions to the choice of MLLM judge.}
We keep the generated videos, metric rubrics, and aggregation protocol fixed,
and replace only the judge used by CAC, CLC, EG, and VLM Alignment. GPT-5 and Gemini-3.1-Pro are judges from outside the Qwen family; 
Qwen3-VL-30B-A3B is the original judge used
in the main evaluation. Higher scores indicate better performance.}
\label{tab:tb6_benchmark_evaluation_robustness}
\small
\setlength{\tabcolsep}{8pt}
\renewcommand{\arraystretch}{1.08}
\begin{tabularx}{\linewidth}{@{}ll
  >{\centering\arraybackslash}X
  >{\centering\arraybackslash}X
  >{\centering\arraybackslash}X
  >{\centering\arraybackslash}X@{}}
\toprule
Method & Judge & CAC $\uparrow$ & CLC $\uparrow$ & EG $\uparrow$ & VLM $\uparrow$ \\
\midrule
\multirow{3}{*}{LongLive}
& GPT-5 & 83.05 & 90.01 & 58.98 & 57.02 \\
& Gemini-3.1-Pro & 82.99 & 89.83 & 55.42 & 51.99 \\
& Qwen3-VL-30B-A3B & 83.06 & 89.97 & 62.40 & 62.22 \\
\midrule
\multirow{3}{*}{MemFlow}
& GPT-5 & 83.90 & 89.81 & 57.67 & 55.30 \\
& Gemini-3.1-Pro & 83.86 & 89.69 & 53.88 & 49.62 \\
& Qwen3-VL-30B-A3B & 83.92 & 89.71 & 61.93 & 60.85 \\
\midrule
\multirow{3}{*}{\textbf{IAMFlow}}
& GPT-5 & 86.09 & 91.37 & 58.54 & 57.53 \\
& Gemini-3.1-Pro & 86.06 & 91.28 & 54.92 & 53.85 \\
& Qwen3-VL-30B-A3B & 86.07 & 91.25 & 62.03 & 62.61 \\
\bottomrule
\end{tabularx}
\end{table}

\subsection{Human Alignment}
To validate whether our evaluation metrics align with human perception, we conduct a user study on 30 samples generated by LongLive, MemFlow, and IAMFlow. We recruit 30 volunteers to rank the three model outputs for each metric, convert the rankings into pairwise preferences, and compute Spearman's rank correlation~\cite{spearman1904} between human preferences and automatic score margins. As shown in Table~\ref{tab:tb7_human_alignment}, NarraStream-Bench metrics are highly consistent with human judgments.
The full human-study protocol is provided in Appendix~\ref{app:human_study}.

\begin{table}[!htbp]
\centering
\caption{\textbf{Spearman's $\rho$ between human preferences and automatic scores.}
These scores show that NarraStream-Bench metrics are highly aligned with human judgments.}
\label{tab:tb7_human_alignment}
\small
\setlength{\tabcolsep}{4.2pt}
\renewcommand{\arraystretch}{0.95}
\setlength{\aboverulesep}{0.25ex}
\setlength{\belowrulesep}{0.35ex}
\resizebox{\linewidth}{!}{%
\begin{tabular}{c|ccccc|ccc|ccc}
\toprule
 & \multicolumn{5}{c|}{Visual Quality} & \multicolumn{3}{c|}{Temporal Consistency} & \multicolumn{3}{c}{ Instruction Compliance} \\
\cmidrule(lr){2-6}\cmidrule(lr){7-9}\cmidrule(lr){10-12}
 & SC & BC & TF & MS & VTSS & BS & CAC & CLC & EG & DT & VLM \\
\midrule
$\rho$ & 0.8936 & 0.9103 & 0.8842 & 0.8774 & 0.8985 & 0.8210 & 0.8316 & 0.8834 & 0.8104 & 0.9053 & 0.8671 \\
\bottomrule
\end{tabular}%
}
\end{table}
\vspace{-8pt}

\section{Conclusion}
We present \textbf{IAMFlow}, a training-free framework for narrative streaming video generation that reframes long-term consistency as identity-aware memory preservation rather than generic context extension.
By organizing historical information around persistent entity IDs and visual attributes, IAMFlow enables streaming video generation to preserve identities across evolving multi-prompt narrative scenarios, where prior memory designs often suffer from identity drift, character duplication, or attribute loss.
To make the proposed framework computationally practical, IAMFlow further integrates a systematic inference acceleration pipeline, including asynchronous visual verification, adaptive prompt transition, and model quantization.
We also introduce \textbf{NarraStream-Bench}, a modern benchmark for multi-prompt narrative streaming video generation.
Extensive experiments show that IAMFlow achieves the best overall performance on NarraStream-Bench.

\noindent\textbf{Limitations and Future Work.}
IAMFlow prioritizes persistent identity consistency, which can make it conservative when prompts demand abrupt semantic changes, complex actions, or fine-grained relational updates.
This reflects a broader trade-off between memory stability and prompt responsiveness in narrative streaming video generation.
Future work will explore adaptive memory and transition strategies that maintain this balance across prompt switches, as well as more flexible entity representations that can capture evolving attributes and relationships without losing identity coherence.

\bibliography{april_aigc}

@article{wan21,
  title={Wan: Open and advanced large-scale video generative models},
  author={Wan, Team and Wang, Ang and Ai, Baole and Wen, Bin and Mao, Chaojie and Xie, Chen-Wei and Chen, Di and Yu, Feiwu and Zhao, Haiming and Yang, Jianxiao and others},
  journal={arXiv preprint arXiv:2503.20314},
  year={2025}
}

@inproceedings{cogvideox,
  title={CogVideoX: Text-to-Video Diffusion Models with An Expert Transformer},
  author={Yang, Zhuoyi and Teng, Jiayan and Zheng, Wendi and Ding, Ming and Huang, Shiyu and Xu, Jiazheng and Yang, Yuanming and Hong, Wenyi and Zhang, Xiaohan and Feng, Guanyu and others},
  booktitle={The Thirteenth International Conference on Learning Representations},
  year={2025}
}

@inproceedings{pyramidalflow,
  title={Pyramidal Flow Matching for Efficient Video Generative Modeling},
  author={Jin, Yang and Sun, Zhicheng and Li, Ningyuan and Xu, Kun and Jiang, Hao and Zhuang, Nan and Huang, Quzhe and Song, Yang and MU, Yadong and Lin, Zhouchen},
  booktitle={The Thirteenth International Conference on Learning Representations},
  year={2025}
}

@inproceedings{nova,
  title={Autoregressive Video Generation without Vector Quantization},
  author={Deng, Haoge and Pan, Ting and Diao, Haiwen and Luo, Zhengxiong and Cui, Yufeng and Lu, Huchuan and Shan, Shiguang and Qi, Yonggang and Wang, Xinlong},
  booktitle={The Thirteenth International Conference on Learning Representations},
  year={2025}
}

@article{diffusionforcing,
  title={Diffusion forcing: Next-token prediction meets full-sequence diffusion},
  author={Chen, Boyuan and Mart{\'\i} Mons{\'o}, Diego and Du, Yilun and Simchowitz, Max and Tedrake, Russ and Sitzmann, Vincent},
  journal={Advances in Neural Information Processing Systems},
  volume={37},
  pages={24081--24125},
  year={2024}
}

@inproceedings{causvid,
  title={From slow bidirectional to fast autoregressive video diffusion models},
  author={Yin, Tianwei and Zhang, Qiang and Zhang, Richard and Freeman, William T and Durand, Fredo and Shechtman, Eli and Huang, Xun},
  booktitle={Proceedings of the IEEE/CVF Conference on Computer Vision and Pattern Recognition},
  pages={22963--22974},
  year={2025}
}

@article{selfforcing,
  title={Self forcing: Bridging the train-test gap in autoregressive video diffusion},
  author={Huang, Xun and Li, Zhengqi and He, Guande and Zhou, Mingyuan and Shechtman, Eli},
  journal={arXiv preprint arXiv:2506.08009},
  year={2025}
}

@article{rollingforcing,
  title={Rolling forcing: Autoregressive long video diffusion in real time},
  author={Liu, Kunhao and Hu, Wenbo and Xu, Jiale and Shan, Ying and Lu, Shijian},
  journal={arXiv preprint arXiv:2509.25161},
  year={2025}
}

@inproceedings{selfforcingpp,
  title={Self-Forcing++: Towards Minute-Scale High-Quality Video Generation},
  author={Cui, Justin and Wu, Jie and Li, Ming and Yang, Tao and Li, Xiaojie and Wang, Rui and Bai, Andrew and Ban, Yuanhao and Hsieh, Cho-Jui},
  booktitle={The Fourteenth International Conference on Learning Representations},
  year={2026}
}

@article{contextforcing,
  title={Context Forcing: Consistent Autoregressive Video Generation with Long Context},
  author={Chen, Shuo and Wei, Cong and Sun, Sun and Nie, Ping and Zhou, Kai and Zhang, Ge and Yang, Ming-Hsuan and Chen, Wenhu},
  journal={arXiv preprint arXiv:2602.06028},
  year={2026}
}

@inproceedings{pavdm,
  title={Progressive autoregressive video diffusion models},
  author={Xie, Desai and Xu, Zhan and Hong, Yicong and Tan, Hao and Liu, Difan and Liu, Feng and Kaufman, Arie and Zhou, Yang},
  booktitle={Proceedings of the Computer Vision and Pattern Recognition Conference},
  pages={6322--6332},
  year={2025}
}

@inproceedings{streamingt2v,
  title={Streamingt2v: Consistent, dynamic, and extendable long video generation from text},
  author={Henschel, Roberto and Khachatryan, Levon and Poghosyan, Hayk and Hayrapetyan, Daniil and Tadevosyan, Vahram and Wang, Zhangyang and Navasardyan, Shant and Shi, Humphrey},
  booktitle={Proceedings of the Computer Vision and Pattern Recognition Conference},
  pages={2568--2577},
  year={2025}
}

@article{longlive,
  title={Longlive: Real-time interactive long video generation},
  author={Yang, Shuai and Huang, Wei and Chu, Ruihang and Xiao, Yicheng and Zhao, Yuyang and Wang, Xianbang and Li, Muyang and Xie, Enze and Chen, Yingcong and Lu, Yao and others},
  journal={arXiv preprint arXiv:2509.22622},
  year={2025}
}

@article{memflow,
  title={Memflow: Flowing adaptive memory for consistent and efficient long video narratives},
  author={Ji, Sihui and Chen, Xi and Yang, Shuai and Tao, Xin and Wan, Pengfei and Zhao, Hengshuang},
  journal={arXiv preprint arXiv:2512.14699},
  year={2025}
}

@article{rewardforcing,
  title={Reward forcing: Efficient streaming video generation with rewarded distribution matching distillation},
  author={Lu, Yunhong and Zeng, Yanhong and Li, Haobo and Ouyang, Hao and Wang, Qiuyu and Cheng, Ka Leong and Zhu, Jiapeng and Cao, Hengyuan and Zhang, Zhipeng and Zhu, Xing and others},
  journal={arXiv preprint arXiv:2512.04678},
  year={2025}
}

@article{deepforcing,
  title={Deep forcing: Training-free long video generation with deep sink and participative compression},
  author={Yi, Jung and Jang, Wooseok and Cho, Paul Hyunbin and Nam, Jisu and Yoon, Heeji and Kim, Seungryong},
  journal={arXiv preprint arXiv:2512.05081},
  year={2025}
}

@article{skyreelsv2,
  title={Skyreels-v2: Infinite-length film generative model},
  author={Chen, Guibin and Lin, Dixuan and Yang, Jiangping and Lin, Chunze and Zhu, Junchen and Fan, Mingyuan and Zhang, Hao and Chen, Sheng and Chen, Zheng and Ma, Chengcheng and others},
  journal={arXiv preprint arXiv:2504.13074},
  year={2025}
}

@article{magi1,
  title={Magi-1: Autoregressive video generation at scale},
  author={Teng, Hansi and Jia, Hongyu and Sun, Lei and Li, Lingzhi and Li, Maolin and Tang, Mingqiu and Han, Shuai and Zhang, Tianning and Zhang, WQ and Luo, Weifeng and others},
  journal={arXiv preprint arXiv:2505.13211},
  year={2025}
}

@article{framepack,
  author= {Lvmin Zhang and Maneesh Agrawala},
  title= {Packing Input Frame Context in Next-Frame Prediction Models for Video Generation},
  journal= {CoRR},
  volume = {abs/2504.12626},
  year = {2025}
}

@article{far,
  title={Long-context autoregressive video modeling with next-frame prediction},
  author={Gu, Yuchao and Mao, Weijia and Shou, Mike Zheng},
  journal={arXiv preprint arXiv:2503.19325},
  year={2025}
}

@article{lovic,
  title={Lovic: Efficient long video generation with context compression},
  author={Jiang, Jiaxiu and Li, Wenbo and Ren, Jingjing and Qiu, Yuping and Guo, Yong and Xu, Xiaogang and Wu, Han and Zuo, Wangmeng},
  journal={arXiv preprint arXiv:2507.12952},
  year={2025}
}

@article{pfp,
  title={Pretraining Frame Preservation in Autoregressive Video Memory Compression},
  author={Zhang, Lvmin and Cai, Shengqu and Li, Muyang and Zeng, Chong and Lu, Beijia and Rao, Anyi and Han, Song and Wetzstein, Gordon and Agrawala, Maneesh},
  journal={arXiv preprint arXiv:2512.23851},
  year={2025}
}

@article{pfvg,
  title={Pack and force your memory: Long-form and consistent video generation},
  author={Wu, Xiaofei and Zhang, Guozhen and Xu, Zhiyong and Zhou, Yuan and Lu, Qinglin and He, Xuming},
  journal={arXiv preprint arXiv:2510.01784},
  year={2025}
}

@article{inflvg,
  title={Inflvg: Reinforce inference-time consistent long video generation with grpo},
  author={Fang, Xueji and Ma, Liyuan and Chen, Zhiyang and Zhou, Mingyuan and Qi, Guo-jun},
  journal={arXiv preprint arXiv:2505.17574},
  year={2025}
}

@article{videossm,
  title={Videossm: Autoregressive long video generation with hybrid state-space memory},
  author={Yu, Yifei and Wu, Xiaoshan and Hu, Xinting and Hu, Tao and Sun, Yangtian and Lyu, Xiaoyang and Wang, Bo and Ma, Lin and Ma, Yuewen and Wang, Zhongrui and others},
  journal={arXiv preprint arXiv:2512.04519},
  year={2025}
}

@inproceedings{ttt-video,
  title={One-minute video generation with test-time training},
  author={Dalal, Karan and Koceja, Daniel and Xu, Jiarui and Zhao, Yue and Han, Shihao and Cheung, Ka Chun and Kautz, Jan and Choi, Yejin and Sun, Yu and Wang, Xiaolong},
  booktitle={Proceedings of the Computer Vision and Pattern Recognition Conference},
  pages={17702--17711},
  year={2025}
}

@inproceedings{slowfastvgen,
  title={SlowFast-VGen: Slow-Fast Learning for Action-Driven Long Video Generation},
  author={Hong, Yining and Liu, Beide and Wu, Maxine and Zhai, Yuanhao and Chang, Kai-Wei and Li, Linjie and Lin, Kevin and Lin, Chung-Ching and Wang, Jianfeng and Yang, Zhengyuan and others},
  booktitle={The Thirteenth International Conference on Learning Representations},
  year={2025}
}

@inproceedings{vllm,
  title={Efficient memory management for large language model serving with pagedattention},
  author={Kwon, Woosuk and Li, Zhuohan and Zhuang, Siyuan and Sheng, Ying and Zheng, Lianmin and Yu, Cody Hao and Gonzalez, Joseph and Zhang, Hao and Stoica, Ion},
  booktitle={Proceedings of the 29th symposium on operating systems principles},
  pages={611--626},
  year={2023}
}

@misc{lightx2v,
  author={LightX2V Contributors},
  title={LightX2V: Light Video Generation Inference Framework},
  year={2025},
  publisher={GitHub},
  journal={GitHub repository},
  howpublished={\url{https://github.com/ModelTC/LightX2V}}
}

@misc{taehv,
  author={Ollin Boer Bohan},
  title={Seraena: WIP Pytorch code for stably training single-step, mode-dropping, deterministic autoencoders},
  year={2024},
  publisher={GitHub},
  journal={GitHub repository},
  howpublished={\url{https://github.com/madebyollin/seraena}},
}

@article{qwen3,
  title={Qwen3 technical report},
  author={Yang, An and Li, Anfeng and Yang, Baosong and Zhang, Beichen and Hui, Binyuan and Zheng, Bo and Yu, Bowen and Gao, Chang and Huang, Chengen and Lv, Chenxu and others},
  journal={arXiv preprint arXiv:2505.09388},
  year={2025}
}

@article{blockvid,
  title={BlockVid: Block Diffusion for High-Quality and Consistent Minute-Long Video Generation},
  author={Zhang, Zeyu and Chang, Shuning and He, Yuanyu and Han, Yizeng and Tang, Jiasheng and Wang, Fan and Zhuang, Bohan},
  journal={arXiv preprint arXiv:2511.22973},
  year={2025}
}

@article{vbenchpp,
  title={Vbench++: Comprehensive and versatile benchmark suite for video generative models},
  author={Huang, Ziqi and Zhang, Fan and Xu, Xiaojie and He, Yinan and Yu, Jiashuo and Dong, Ziyue and Ma, Qianli and Chanpaisit, Nattapol and Si, Chenyang and Jiang, Yuming and others},
  journal={IEEE Transactions on Pattern Analysis and Machine Intelligence},
  year={2025},
  publisher={IEEE}
}

@inproceedings{dit,
  title={Scalable diffusion models with transformers},
  author={Peebles, William and Xie, Saining},
  booktitle={Proceedings of the IEEE/CVF international conference on computer vision},
  pages={4195--4205},
  year={2023}
}

@article{attention,
  title={Attention is all you need},
  author={Vaswani, Ashish and Shazeer, Noam and Parmar, Niki and Uszkoreit, Jakob and Jones, Llion and Gomez, Aidan N and Kaiser, {\L}ukasz and Polosukhin, Illia},
  journal={Advances in neural information processing systems},
  volume={30},
  year={2017}
}

@article{qwen3vl,
  title={Qwen3-vl technical report},
  author={Bai, Shuai and Cai, Yuxuan and Chen, Ruizhe and Chen, Keqin and Chen, Xionghui and Cheng, Zesen and Deng, Lianghao and Ding, Wei and Gao, Chang and Ge, Chunjiang and others},
  journal={arXiv preprint arXiv:2511.21631},
  year={2025}
}

@inproceedings{vbench,
  title={Vbench: Comprehensive benchmark suite for video generative models},
  author={Huang, Ziqi and He, Yinan and Yu, Jiashuo and Zhang, Fan and Si, Chenyang and Jiang, Yuming and Zhang, Yuanhan and Wu, Tianxing and Jin, Qingyang and Chanpaisit, Nattapol and others},
  booktitle={Proceedings of the IEEE/CVF Conference on Computer Vision and Pattern Recognition},
  pages={21807--21818},
  year={2024}
}

@inproceedings{videoscore,
  title={Videoscore: Building automatic metrics to simulate fine-grained human feedback for video generation},
  author={He, Xuan and Jiang, Dongfu and Zhang, Ge and Ku, Max and Soni, Achint and Siu, Sherman and Chen, Haonan and Chandra, Abhranil and Jiang, Ziyan and Arulraj, Aaran and others},
  booktitle={Proceedings of the 2024 Conference on Empirical Methods in Natural Language Processing},
  pages={2105--2123},
  year={2024}
}

@inproceedings{t2vcompbench,
  title={T2v-compbench: A comprehensive benchmark for compositional text-to-video generation},
  author={Sun, Kaiyue and Huang, Kaiyi and Liu, Xian and Wu, Yue and Xu, Zihan and Li, Zhenguo and Liu, Xihui},
  booktitle={Proceedings of the Computer Vision and Pattern Recognition Conference},
  pages={8406--8416},
  year={2025}
}

@inproceedings{dino,
  title={Emerging properties in self-supervised vision transformers},
  author={Caron, Mathilde and Touvron, Hugo and Misra, Ishan and J{\'e}gou, Herv{\'e} and Mairal, Julien and Bojanowski, Piotr and Joulin, Armand},
  booktitle={Proceedings of the IEEE/CVF international conference on computer vision},
  pages={9650--9660},
  year={2021}
}

@inproceedings{clip,
  title={Learning transferable visual models from natural language supervision},
  author={Radford, Alec and Kim, Jong Wook and Hallacy, Chris and Ramesh, Aditya and Goh, Gabriel and Agarwal, Sandhini and Sastry, Girish and Askell, Amanda and Mishkin, Pamela and Clark, Jack and others},
  booktitle={International conference on machine learning},
  pages={8748--8763},
  year={2021},
  organization={PmLR}
}

@inproceedings{raft,
  title={Raft: Recurrent all-pairs field transforms for optical flow},
  author={Teed, Zachary and Deng, Jia},
  booktitle={European conference on computer vision},
  pages={402--419},
  year={2020},
  organization={Springer}
}

@inproceedings{amt,
  title={Amt: All-pairs multi-field transforms for efficient frame interpolation},
  author={Li, Zhen and Zhu, Zuo-Liang and Han, Ling-Hao and Hou, Qibin and Guo, Chun-Le and Cheng, Ming-Ming},
  booktitle={Proceedings of the IEEE/CVF Conference on Computer Vision and Pattern Recognition},
  pages={9801--9810},
  year={2023}
}

@inproceedings{languagebind,
  title={Languagebind: Extending video-language pretraining to n-modality by language-based semantic alignment},
  author={Zhu, Bin and Lin, Bin and Ning, Munan and Yan, Yang and Cui, Jiaxi and HongFa, WANG and Pang, Yatian and Jiang, Wenhao and Zhang, Junwu and Li, Zongwei and others},
  booktitle={The Twelfth International Conference on Learning Representations},
  year={2024}
}

@inproceedings{koala36m,
  title={Koala-36m: A large-scale video dataset improving consistency between fine-grained conditions and video content},
  author={Wang, Qiuheng and Shi, Yukai and Ou, Jiarong and Chen, Rui and Lin, Ke and Wang, Jiahao and Jiang, Boyuan and Yang, Haotian and Zheng, Mingwu and Tao, Xin and others},
  booktitle={Proceedings of the Computer Vision and Pattern Recognition Conference},
  pages={8428--8437},
  year={2025}
}

@article{ivebench,
  title={Ivebench: Modern benchmark suite for instruction-guided video editing assessment},
  author={Chen, Yinan and Zhang, Jiangning and Hu, Teng and Zeng, Yuxiang and Xue, Zhucun and He, Qingdong and Wang, Chengjie and Liu, Yong and Hu, Xiaobin and Yan, Shuicheng},
  journal={arXiv preprint arXiv:2510.11647},
  year={2025}
}

@inproceedings{moviebench,
  title={Moviebench: A hierarchical movie level dataset for long video generation},
  author={Wu, Weijia and Liu, Mingyu and Zhu, Zeyu and Xia, Xi and Feng, Haoen and Wang, Wen and Lin, Kevin Qinghong and Shen, Chunhua and Shou, Mike Zheng},
  booktitle={Proceedings of the Computer Vision and Pattern Recognition Conference},
  pages={28984--28994},
  year={2025}
}

@article{narrlv,
  title={NarrLV: Towards a Comprehensive Narrative-Centric Evaluation for Long Video Generation},
  author={Feng, Xiaokun and Yu, Haiming and Wu, Meiqi and Hu, Shiyu and Chen, Jintao and Zhu, Chen and Wu, Jiahong and Chu, Xiangxiang and Huang, Kaiqi},
  journal={arXiv preprint arXiv:2507.11245},
  year={2025}
}

@inproceedings{longvideosurvey,
  title={A survey on long-video storytelling generation: architectures, consistency, and cinematic quality},
  author={Elmoghany, Mohamed and Rossi, Ryan and Yoon, Seunghyun and Mukherjee, Subhojyoti and Bakr, Eslam Mohamed and Mathur, Puneet and Wu, Gang and Lai, Viet Dac and Lipka, Nedim and Zhang, Ruiyi and others},
  booktitle={Proceedings of the IEEE/CVF International Conference on Computer Vision},
  pages={7023--7035},
  year={2025}
}

@article{quantvideogen,
  title={Quant VideoGen: Auto-Regressive Long Video Generation via 2-Bit KV-Cache Quantization},
  author={Xi, Haocheng and Yang, Shuo and Zhao, Yilong and Li, Muyang and Cai, Han and Li, Xingyang and Lin, Yujun and Zhang, Zhuoyang and Zhang, Jintao and Li, Xiuyu and others},
  journal={arXiv preprint arXiv:2602.02958},
  year={2026}
}

@article{fp8formats,
  title={Fp8 formats for deep learning},
  author={Micikevicius, Paulius and Stosic, Dusan and Burgess, Neil and Cornea, Marius and Dubey, Pradeep and Grisenthwaite, Richard and Ha, Sangwon and Heinecke, Alexander and Judd, Patrick and Kamalu, John and others},
  journal={arXiv preprint arXiv:2209.05433},
  year={2022}
}

@article{spearman1904,
  title={The Proof and Measurement of Association between Two Things},
  author={Spearman, C},
  journal={The American Journal of Psychology},
  volume={15},
  number={1},
  pages={72--101},
  year={1904}
}

@article{streamingllm,
  title={Efficient streaming language models with attention sinks},
  author={Xiao, Guangxuan and Tian, Yuandong and Chen, Beidi and Han, Song and Lewis, Mike},
  journal={arXiv preprint arXiv:2309.17453},
  year={2023}
}

\newpage
\appendix
\onecolumn
\renewcommand{\thesection}{\Alph{section}}
\setcounter{section}{0}

{\Large\bfseries Appendix\par}

\vspace{1.0em}

{\bfseries Contents}

\vspace{0.6em}

\begingroup
\renewcommand{\arraystretch}{1.5}
\noindent\begin{tabular*}{\linewidth}{@{}p{1.8em}@{}p{0.72\linewidth}@{\extracolsep{\fill}}r@{}}
\hyperref[app:memory_implement]{\textbf{A}} & \hyperref[app:memory_implement]{\textbf{Identity-Aware Memory Implementation Details}} & \pageref{app:memory_implement}\\
\hyperref[app:inference_acceleration]{\textbf{B}} & \hyperref[app:inference_acceleration]{\textbf{Systematic Inference Acceleration Details}} & \pageref{app:inference_acceleration}\\
\hyperref[app:agent_prompts]{\textbf{C}} & \hyperref[app:agent_prompts]{\textbf{LLM/VLM Prompt Templates}} & \pageref{app:agent_prompts}\\
\hyperref[app:narrstream_construction]{\textbf{D}} & \hyperref[app:narrstream_construction]{\textbf{NarraStream-Bench Database Construction}} & \pageref{app:narrstream_construction}\\
\hyperref[app:narrstream_metrics]{\textbf{E}} & \hyperref[app:narrstream_metrics]{\textbf{Metric Details in NarraStream-Bench}} & \pageref{app:narrstream_metrics}\\
\hyperref[app:baseline_implementation]{\textbf{F}} & \hyperref[app:baseline_implementation]{\textbf{Baseline Implementation Details}} & \pageref{app:baseline_implementation}\\
\hyperref[app:human_study]{\textbf{G}} & \hyperref[app:human_study]{\textbf{Human Alignment Study Details}} & \pageref{app:human_study}\\
\hyperref[app:limitations]{\textbf{H}} & \hyperref[app:limitations]{\textbf{Limitation Analysis}} & \pageref{app:limitations}\\
\hyperref[app:more_qual_details]{\textbf{I}} & \hyperref[app:more_qual_details]{\textbf{More Qualitative Results and Details}} & \pageref{app:more_qual_details}\\
\hyperref[app:ethics]{\textbf{J}} & \hyperref[app:ethics]{\textbf{Ethics Statement}} & \pageref{app:ethics}\\
\hyperref[app:reproduce]{\textbf{K}} & \hyperref[app:reproduce]{\textbf{Reproducibility Statement}} & \pageref{app:reproduce}\\
\hyperref[app:social]{\textbf{L}} & \hyperref[app:social]{\textbf{Broader Social Impact}} & \pageref{app:social}
\end{tabular*}
\endgroup

\section{Identity-Aware Memory Implementation Details}
\label{app:memory_implement}

This appendix supplements Sec.~\ref{sec:method_id_memory} with the implementation of the identity-aware memory module.

\paragraph{Runtime memory state.}
IAMFlow maintains three data structures during generation.
The global entity registry $\mathcal{R}$ maps each persistent ID to a canonical name, a list of textual aliases observed across prompts, a merged attribute list, and per-prompt instances.
The frame archive $\mathcal{F}$ stores selected historical frames with their prompt index, associated entity IDs, entity score, VLM visual score, fused score, and per-layer KV cache.
The active identity memory $m^{\mathrm{id}}$ is a compact list of archived frame IDs.
The selected KV tensors are sorted by temporal order, concatenated block-wise, and written into the model memory bank before denoising.
\begin{paperalgorithm}{IAMFlow Streaming Generation with Identity-Aware Memory}
\label{alg:iamflow_streaming}
\noindent\textbf{Require:} Causal video generator $G_\theta$, prompt sequence $\{p_t\}_{t=1}^{T}$, local window size $W_{\mathrm{loc}}$\\
\textbf{Require:} Identity-memory budget $B$, asynchronous verification worker\\[0.35ex]
\begin{tabularx}{\linewidth}{@{}r@{\quad}X@{}}
1: & Initialize global registry $\mathcal{R}\leftarrow\emptyset$, frame archive $\mathcal{F}\leftarrow\emptyset$, local cache $\mathcal{L}\leftarrow\emptyset$\\
2: & \textbf{for} prompt index $t=1,\ldots,T$ \textbf{do}\\
3: & \hspace{1.2em}Extract entities and attributes $\mathcal{E}_t\leftarrow\textsc{LLMParse}(p_t)$\\
4: & \hspace{1.2em}Assign persistent IDs and update registry $(\mathcal{G}_t,\mathcal{R})\leftarrow\textsc{MatchOrCreateID}(\mathcal{E}_t,\mathcal{R})$\\
5: & \hspace{1.2em}$m^{\mathrm{id}}_t\leftarrow\textsc{GreedyIDRetrieve}(\mathcal{F},\mathcal{G}_t,B)$\\
6: & \hspace{1.2em}Inject the KV cache of $m^{\mathrm{id}}_t$ into the model memory bank\\
7: & \hspace{1.2em}\textbf{for} chunk index $n=1,2,\ldots$ under prompt $p_t$ \textbf{do}\\
8: & \hspace{2.4em}$z_n\leftarrow\textsc{DenoiseChunk}(G_\theta,p_t,\mathcal{L},m^{\mathrm{id}}_t)$\\
9: & \hspace{2.4em}Submit $z_n$ to the asynchronous worker for visual scoring; request attribute check if $n=1$\\
10: & \hspace{2.4em}Append $z_n$ to $\mathcal{L}$ and update the clean-context KV cache\\
11: & \hspace{2.4em}\textbf{if} the local-window update evicts a chunk \textbf{then}\\
12: & \hspace{3.6em}$c\leftarrow$ the evicted chunk\\
13: & \hspace{2.4em}\textbf{else}\\
14: & \hspace{3.6em}$c\leftarrow$ the clean-context cache after the context update\\
15: & \hspace{2.4em}\textbf{end if}\\
16: & \hspace{2.4em}Collect worker results needed by $c$; if first-chunk verification returns, correct $\mathcal{R}$\\
17: & \hspace{2.4em}Build entity-token weights from active IDs $\mathcal{G}_t$ and registry $\mathcal{R}$\\
18: & \hspace{2.4em}Score frames in $c$ using $s(f)=(1-\lambda)\hat{s}_{\mathrm{entity}}(f)+\lambda s_{\mathrm{visual}}(f)$\\
19: & \hspace{2.4em}Archive the top frame and its entity IDs, fused score, and KV cache into $\mathcal{F}$\\
20: & \hspace{1.2em}\textbf{end for}\\
21: & \textbf{end for}
\end{tabularx}
\end{paperalgorithm}

\paragraph{Ablation fallback protocols.}
For the memory-mechanism ablations in Table~\ref{tab:tb4_ablation_combined}, all unablated components are kept unchanged so that each row isolates a single design choice.
In \textit{w/o LLM planning}, we disable both LLM-based entity extraction and LLM-based global ID matching.
The system falls back to a deterministic text heuristic: human-related noun phrases and explicitly named characters are extracted from the segment prompt, normalized by lowercasing and stripping articles, and linked across prompts only by exact normalized surface-form matches.
Unmatched entities are assigned new segment-level IDs, so this variant does not perform semantic alias resolution, pronoun resolution, or attribute-based identity matching, while the frame archive, identity-aware scoring, retrieval, VLM verification, and prompt transition remain active.
In \textit{w/o VLM verification}, we disable asynchronous VLM scoring and attribute correction; archived frames are ranked only by the normalized entity score $\hat{s}_{\mathrm{entity}}(f)$, and the registry is updated only from prompt-side observations.
In \textit{w/o memory bank}, we remove the global registry $\mathcal{R}$, frame archive $\mathcal{F}$, active identity memory $m^{\mathrm{id}}$, and all long-term identity-memory retrieval and KV injection.
Generation therefore relies only on the sink frames and sliding local window of the base autoregressive generator under the same prompt schedule and transition policy.

\paragraph{Prompt-level entity processing.}
At the first chunk of each prompt, Qwen3-4B-Instruct extracts human entities and stable visual attributes from the prompt.
For the first prompt, all extracted entities are assigned new IDs.
For later prompts, explicit novelty markers such as ``another'', ``new'', or ``different'' trigger allocation of a new ID; otherwise, the LLM compares the current entity descriptor with the existing registry and either reuses a matched ID or allocates a new one.
When an ID is reused, IAMFlow appends the new alias and prompt instance to $\mathcal{R}$ and merges newly observed attributes.
This converts prompt-level references such as ``young man'', ``the protagonist'', and ``he'' into a shared identity address.

\paragraph{Frame archival and identity-aware scoring.}
After each generated chunk, IAMFlow archives one representative latent frame.
For early chunks before the local cache rolls, the frame is selected from the clean-context KV cache after the context update; afterwards, it is selected from the chunk leaving the local window.
For the active prompt, IAMFlow builds an entity-token weight vector $\omega\in\mathbb{R}^{S}$ with a fixed heuristic.
All raw token weights are initialized to $1.0$.
For each active entity name and extracted attribute, we locate its lowercase character span in the prompt, map the character-span ratio to token indices, and expand the span by $0.02S$ tokens on both sides.
Tokens in these matched spans are assigned weight $2.5$.
Unmatched tokens in the first $8\%$ and last $8\%$ of the prompt are assigned weights $0.7$ and $0.5$, respectively, while other unmatched tokens remain at $1.0$.
If no entity or attribute span is matched, tokens from $10\%$ to $85\%$ of the prompt length receive weight $1.5$ and all other tokens remain at $1.0$; if no entity is detected, all tokens receive weight $1.0$.
The normalized weights used below are
\begin{equation}
\tilde{\omega}_{u}=\frac{\omega_u}{\sum_{v=1}^{S}\omega_v+10^{-8}} .
\end{equation}
Given cached text keys $K^{\mathrm{text}}$ and candidate visual keys $K^{\mathrm{vis}}$, the implementation aggregates the weighted text-key anchor and frame key as
\begin{equation}
\bar{\mathbf{r}}_{\mathrm{id},h}=\sum_{u}\tilde{\omega}_{u}\mathbf{K}^{\mathrm{text}}_{u,h},
\quad
\bar{\mathbf{k}}_{f,h}=\frac{1}{n_f}\sum_{v=1}^{n_f}\mathbf{K}^{\mathrm{vis}}_{f,v,h},
\end{equation}
and scores each candidate frame by
\begin{equation}
s_{\mathrm{entity}}(f)=
\frac{1}{H}\sum_{h=1}^{H}
\frac{\langle \bar{\mathbf{r}}_{\mathrm{id},h},\bar{\mathbf{k}}_{f,h}\rangle}{\sqrt{d}} .
\end{equation}
We use a representative early transformer layer in implementation. This keeps frame ranking lightweight while exploiting the generator's internal text-conditioned key-space compatibility.
This corresponds to the single-layer implementation of Eq.~\ref{eq:entity_score}, with $\mathcal{L}=\{l^\star\}$ and $\beta_{l^\star}=1$.
The entity score is min-max normalized within the chunk and fused with the VLM score as
\begin{equation}
s(f)=(1-\lambda)\hat{s}_{\mathrm{entity}}(f)+\lambda s_{\mathrm{visual}}(f),
\end{equation}
where $\lambda=0.3$.
The highest-scoring frame is stored in $\mathcal{F}$ together with its associated entity IDs and KV cache.

\paragraph{Dynamic retrieval and memory injection.}
At each prompt switch, IAMFlow retrieves archived frames for the active entity IDs using a greedy set-cover strategy.
The dynamic budget is the number of frames needed to cover the required IDs, capped by the maximum identity-memory size.
When multiple entities are active and enough archive candidates exist, IAMFlow keeps at least two memory frames to avoid collapsing different identities into a single reference.
At each greedy step, the selected frame maximizes the number of still-uncovered IDs, with entity score used as the tie-breaker.
If the budget is not exhausted after all IDs are covered, remaining slots are filled by high-scoring archived frames.
The selected KV caches are concatenated and injected into the memory bank; the sink memory and local sliding window remain unchanged.

\paragraph{Asynchronous VLM verification.}
IAMFlow decodes generated chunks with a background VAE worker and submits sampled frames to Qwen3-VL-2B-Instruct for visual verification.
The VLM scores three sampled frames from each chunk and returns $s_{\mathrm{visual}}(f)\in[0,1]$ for the fusion above.
For the first chunk of each prompt, the same VLM call also verifies the registry attributes against rendered pixels and writes corrected attribute lists back to $\mathcal{R}$.
The background worker preserves causal VAE order, while the main DiT stream continues denoising the next chunk.
In practice, the three-chunk eviction lag hides most VAE and VLM latency, so visual verification improves memory quality without adding a blocking step to the common generation path.

\section{Systematic Inference Acceleration Details}
\label{app:inference_acceleration}

\begin{figure}[t]
  \centering
  \includegraphics[width=\linewidth]{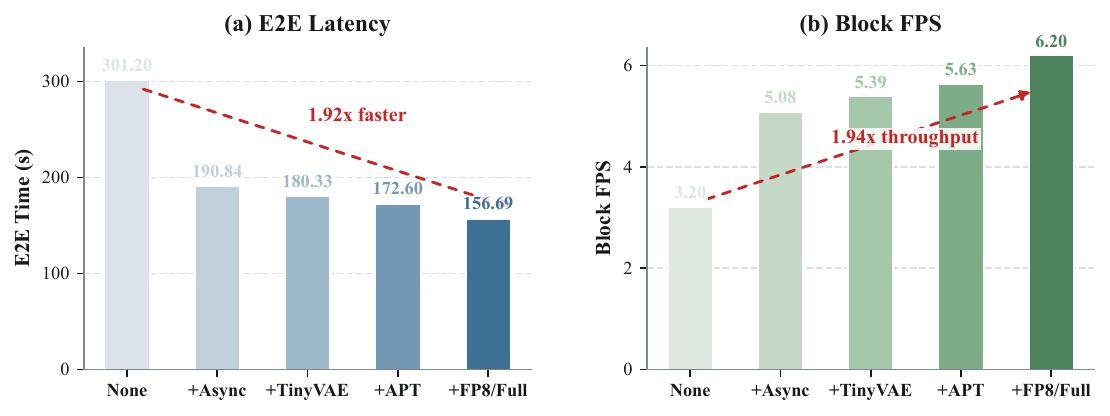}
  \caption{
  \textbf{Build-up analysis of IAMFlow systematic inference acceleration under NVIDIA H20 GPUs}.
  Compared with IAMFlow without systematic inference acceleration, the full accelerated IAMFlow substantially reduces end-to-end latency and improves block-level throughput. The build-up starts from the identity-aware memory pipeline without acceleration and progressively adds the Asynchronous Verification Pipeline, TinyVAE decoding, Adaptive Prompt Transition, and FP8 DiT execution.}
  \label{fig:app_efficiency_build_up}
\end{figure}

This appendix gives the implementation details behind the systematic inference acceleration pipeline summarized in Sec.~\ref{sec:method_acceleration}.
The main design goal is to keep identity-aware memory off the critical generation path: prompt understanding runs only at segment boundaries, visual verification overlaps with DiT denoising, and prompt switches avoid full temporal recaching.
Fig.~\ref{fig:app_efficiency_build_up} shows that, compared with IAMFlow without systematic inference acceleration, the full accelerated IAMFlow achieves substantial end-to-end speedup and higher block-level throughput.

\paragraph{Adaptive prompt transition.}
Autoregressive multi-prompt generation usually refreshes the temporal cache at each prompt switch by rerunning the recent latent context through the DiT under the new text condition, so that cross-attention states match the new prompt.
This recache pass scales with the local window length and reruns much of the denoising model.
IAMFlow replaces it with Adaptive Prompt Transition (APT): before switching prompts, the current cross-attention keys and values are copied into a preallocated buffer; the new prompt cache is lazily initialized on the next forward pass; each cross-attention layer then uses
\begin{equation}
  \mathbf{K}_{\tau}=(1-\alpha_{\tau})\mathbf{K}_{\mathrm{old}}+\alpha_{\tau}\mathbf{K}_{\mathrm{new}},
  \quad
  \mathbf{V}_{\tau}=(1-\alpha_{\tau})\mathbf{V}_{\mathrm{old}}+\alpha_{\tau}\mathbf{V}_{\mathrm{new}} .
\end{equation}
This blend operates in the cross-attention conditioning space rather than in pixel or noise space.
We use it as a local interpolation between old and new semantic conditioning, so the denoising direction changes continuously after a prompt switch instead of being reset by a hard condition replacement.
Let $\tau$ denote the number of latent frames generated after a prompt switch and let $d_{\mathrm{delay}}=3$.
During the delay phase, $\alpha_{\tau}=0$.
Afterwards, $\alpha_{\tau}$ follows a cosine schedule,
\begin{equation}
\alpha_{\tau}=
\frac{1}{2}\left(1-\cos\left(\pi\frac{\tau-d_{\mathrm{delay}}}{W_{\mathrm{apt}}}\right)\right),
\quad
\tau\in[d_{\mathrm{delay}},d_{\mathrm{delay}}+W_{\mathrm{apt}}],
\end{equation}
and becomes 1 after the transition.
The window length is adaptive. Let $\bar{\mathbf{e}}_i$ and
$\bar{\mathbf{e}}_{i+1}$ be the mean-pooled text-encoder embeddings of adjacent
prompts. We use the clipped cosine distance
\[
\delta=\operatorname{clip}\bigl(1-\cos(\bar{\mathbf{e}}_i,\bar{\mathbf{e}}_{i+1}),0,1\bigr)
\]
and set
\begin{equation}
  W_{\mathrm{apt}}=\mathrm{snap}\bigl(W_{\min}+\delta(W_{\max}-W_{\min})\bigr),
  \quad
  W_{\min}=3,\; W_{\max}=15,
\end{equation}
where $\mathrm{snap}(\cdot)$ aligns the window to the 3-frame chunk boundary.
The clipping keeps $W_{\mathrm{apt}}$ within the declared window range before snapping.
Since prompt embeddings are already computed before generation, this adds no extra text-encoder pass.
Small edits therefore use short transitions, while larger scene or role changes receive longer blending windows.

\paragraph{Asynchronous Verification Pipeline.}
IAMFlow needs pixel-space evidence for VLM visual scoring and for first-chunk attribute correction.
If VAE decoding and VLM inference were synchronous, visual verification would directly increase user-facing latency.
We instead use a single background worker for streaming VAE decoding and a separate asynchronous VLM executor.
After each chunk is denoised, the main DiT stream submits the latent chunk to the background worker and immediately continues to cache update and the next denoising step.
The single-worker VAE queue preserves causal decoder order.

The schedule exploits the autoregressive eviction lag.
The local portion of the KV cache consists of 3 sink latent frames and a 9-frame sliding local window, so the local window spans three 3-frame chunks; the retrieved identity-memory block is separate from this eviction schedule.
For chunk $n\geq4$, the frame selected for archival comes from the chunk leaving the sliding local window, i.e., chunk $n-3$.
Thus the VAE and VLM have roughly three chunks of DiT computation to finish before their scores are needed.
For the first three chunks of a prompt, where no local chunk has been evicted, IAMFlow archives from the clean-context cache after cache update, allowing VAE decoding to overlap with the clean-context forward pass.
The VLM request is bounded by sampling three frames from each 12-frame decoded chunk at pixel-frame indices $\{0,4,8\}$ and resizing them to $256{\times}256$.

\paragraph{Efficient deployment.}
The efficiency experiments use a LightX2V-style FP8 checkpoint for the DiT linear layers while keeping cache tensors and numerically sensitive operations in BF16/FP32.
This targets the dominant DiT matrix-multiplication cost without changing the identity-aware memory paradigm.
IAMFlow also separates diffusion and language-model workloads: the DiT, text encoder, and VAE run on the generation device, while the LLM and VLM can be pinned to another device and served through vLLM.
The LLM is invoked once per prompt segment for entity extraction and ID matching, and the VLM runs asynchronously on downsampled frames.
For output decoding, IAMFlow replaces the default Wan VAE with a TinyVAE adapted to the Wan2.1 latent layout, greatly reducing the memory footprint and processing latency; because chunks are already decoded asynchronously during generation, the final output step only concatenates decoded chunks instead of launching another full-video VAE pass.

\paragraph{Measurement protocol.}
All efficiency numbers are measured after model loading and warmup.
All methods are evaluated under the same two-GPU NVIDIA H20 budget.
For baselines without LLM/VLM workloads, the second GPU is used to offload VAE decoding when applicable, so the reported speedups do not come from allocating more hardware to IAMFlow.
End-to-end latency is measured from prompt encoding to the complete pixel-level video output, including prompt encoding, LLM entity processing, memory retrieval and injection, DiT denoising, any blocking wait caused by asynchronous VAE/VLM work, and final pixel-video assembly.
FPS is computed from chunk-level throughput: in our Wan2.1 setting, each latent chunk corresponds to 12 decoded pixel frames, so FPS is calculated as $12$ divided by the average generation time of one chunk.
IAMFlow is faster than prior baselines because its systematic inference acceleration pipeline removes full prompt-switch recaching, overlaps visual verification with denoising, reuses chunk-wise decoded pixels for final output, and reduces the dominant DiT compute with FP8 execution.

\section{LLM/VLM Prompt Templates}
\label{app:agent_prompts}

This appendix lists the prompt templates used by IAMFlow's language and vision agents, together with the structured prompts used to construct NarraStream-Bench. All runtime agent calls use deterministic decoding. The LLM entity extraction and ID matching calls use temperature 0; the VLM verification call uses temperature 0.0. The implementation parses only JSON objects from the assistant response and discards surrounding Markdown or natural-language explanations.

\paragraph{LLM entity extraction.}
At the beginning of each prompt segment, IAMFlow sends the current video prompt as the user message and uses the following instruction to extract identity-bearing human entities. We restrict extraction to human characters because the memory bank assigns persistent IDs only to character identities.

\begin{prompttemplatebox}{LLM Entity Extraction Prompt}
[System message]
Extract human characters from the video prompt.

RULES:
1. "entities": ONLY human/person characters (man, woman,
   protagonist, etc.)
   - Extract ONLY visual/physical attributes: hair, clothing,
     accessories, body type, age, skin, facial features
   - DO NOT extract behavioral states (walking, nodding, reading,
     sitting) or emotions (quiet, contemplative, happy)
   - Keep entity names short
OUTPUT FORMAT (JSON object only, no explanation):
{"entities": [{"entity": "<name>", "attrs": ["<attr1>", "<attr2>"]}]}

If no humans found, entities should be [].

[User message]
<prompt_text>
\end{prompttemplatebox}

The expected payload is
\begin{equation}
\{\texttt{entities}: [\{\texttt{entity}: e,\texttt{attrs}: [a_1,\ldots,a_m]\}]\}.
\end{equation}
The parser first searches for the outer JSON object. If that fails, it accepts the legacy list-only entity format and then falls back to regular-expression extraction of \texttt{entity} and \texttt{attrs} fields. Malformed entities are dropped rather than propagated into the memory bank.

\paragraph{LLM global ID matching.}
For the first prompt, IAMFlow allocates new global IDs to all extracted entities. For later prompts, explicit novelty markers such as ``another'', ``other'', ``new'', ``different'', ``second'', and ``third'' allocate a new ID before calling the LLM. All other entities are matched against the current registry with the prompt below.

\begin{prompttemplatebox}{LLM Global ID Matching Prompt}
[System message]
Match a new character to existing characters.

TASK: Given a new character description and existing character registry,
determine if they refer to the same person.

MATCHING RULES:
1. Words like "protagonist", "main character", "he", "she" usually
   refer to previously introduced characters
2. Matching clothing or appearance attributes indicates the same person
3. Words like "another", "other", "new", "different" indicate a NEW
   person - return null

OUTPUT FORMAT (JSON only, no explanation):
{"matched_id": <number or null>}

[User message]
New character description:
"<entity>: <attr1>, <attr2>, ..."

Existing characters:
ID <gid>: <alias1>/<alias2>: <attr1>, <attr2>, ...

Does the new character match any existing one? If yes, return the ID.
If no, return null.
Output JSON only: {"matched_id": <number or null>}
\end{prompttemplatebox}

The implementation accepts a match only when \texttt{matched\_id} is present and the ID already exists in the registry. Otherwise, IAMFlow allocates a new ID. This conservative fallback prevents an uncertain match from merging two identities and polluting later memory retrieval.

\paragraph{VLM visual scoring and attribute correction.}
After each generated chunk, IAMFlow samples three frames from the decoded pixels, resizes them to $256\times256$, and sends them with the prompt below. The sampled frame indices are 0, 4, and 8 in the 12-frame pixel chunk, corresponding to latent frame indices 0, 1, and 2. The VLM returns frame-level visual scores in $[0,1]$; for the first chunk of a prompt, it also verifies the current global registry attributes.

\begin{prompttemplatebox}{VLM Frame Scoring and Attribute Verification}
You are evaluating video frames for visual quality.
The video prompt is: "<prompt_text>"
Key entities: <entity_1>, <entity_2>, ...

Rate each of the <num_frames> frames from 0.0 to 1.0 on how well they
visually represent the entities and match the prompt description.
Consider clarity, consistency, and visual quality.

Also verify these entity attributes against the frames:
  ID <gid> (<name>): <attr1>, <attr2>, ...
If any attribute is incorrect based on what you see, provide corrections.

Respond ONLY with valid JSON:
{"scores": [<float>, ...], "corrections": null}
If there are attribute corrections, use:
{"scores": [<float>, ...],
 "corrections": {"<global_id>": {"corrected_attrs": ["attr1", "attr2"]}}}
\end{prompttemplatebox}

When the request is not the first chunk of a prompt, the attribute-verification block is omitted and the expected \texttt{corrections} value is \texttt{null}. The parser clips all scores into $[0,1]$, pads missing frame scores with 0.5, and ignores malformed correction entries. If VLM inference or parsing fails, IAMFlow uses a neutral score of 0.5 for every sampled frame and applies no registry correction.

\paragraph{NarraStream-Bench prompt generation.}
For NarraStream-Bench prompt generation, each example is first represented as a structured prompt plan that specifies the scene anchor and setting, character registry, six segment beats, visible and off-screen characters, reference mode, and pose mode. The LLM writes the first segment with a stricter introduction prompt and writes the remaining five segments with a continuity-aware follow-up prompt.

\begin{prompttemplatebox}{NarraStream-Bench First-Segment Prompt Template}
You are generating the first segment prompt for a 60-second single-scene
interactive benchmark video.

Requirements:
- Write exactly one English paragraph.
- Output English only.
- Do not use Chinese characters.
- Do not mix English with any other language.
- Keep the final prompt between 20 and 90 English words.
- Prefer 2-3 short sentences.
- Do not exceed 4 short sentences.
- Keep one consistent scene anchor and background stable.
- You MUST copy the exact SCENE_ANCHOR phrase verbatim in the output.
- Do NOT paraphrase or replace the SCENE_ANCHOR wording.
- Do NOT output control labels such as SCENE_ANCHOR:, SCENE_SETTING:,
  REFERENCE_MODE:, BEAT:, or VISIBLE_CHARACTERS:.
- Avoid dialogue and quotation marks.
- Introduce only the characters that are visible in segment 1.
- Use explicit visual descriptions for visible characters.
- Respect the requested pose mode and do not default to making every
  character sit at a table.
- Keep the wording compatible with LongLive, MemFlow, and IAMFlow:
  restate subject anchors and scene anchors clearly.
- Keep the style close to concise visual narration, not literary prose.
- Output only the prompt text.
\end{prompttemplatebox}

For follow-up segments, the system message changes the identity policy while keeping the same length and formatting constraints.

\begin{prompttemplatebox}{NarraStream-Bench Follow-Up Prompt Template}
You are generating a follow-up segment prompt for a 60-second single-scene
interactive benchmark video.

Requirements:
- Write exactly one English paragraph.
- Output English only.
- Keep the final prompt between 20 and 90 English words.
- Prefer 2-3 short sentences.
- Keep one consistent scene anchor and one consistent venue identity unless
  the beat explicitly specifies a nearby sub-area.
- You MUST copy the exact SCENE_ANCHOR phrase verbatim in the output.
- Do NOT output control labels such as SCENE_ANCHOR:, SCENE_SETTING:,
  REFERENCE_MODE:, BEAT:, or VISIBLE_CHARACTERS:.
- Avoid dialogue and quotation marks.
- Respect the requested reference mode for this segment:
  explicit_description, partial_reference, or ambiguous_reference.
- Do not restate full appearance descriptions for every visible character
  unless identity would become unclear.
- Use at most one brief background reminder sentence.
- Keep visible character identities stable and never invent new appearance
  details for old characters.
- Mention off-screen characters only when needed for entry-exit continuity.
- Respect the requested pose mode and do not collapse every segment into
  the same seated table posture.
- Keep the style concise, visual, and compatible with LongLive, MemFlow,
  and IAMFlow.
- Output only the prompt text.
\end{prompttemplatebox}

The user message supplies the exact scene anchor, scene setting, main challenge, pose guidance, required reference mode, target beat, visible characters, off-screen characters, the full character registry, and all previous prompts. This makes each segment self-contained enough for text-to-video models while still testing long-range reference resolution.

\paragraph{NarraStream-Bench evaluation prompts.}
To make the reported NarraStream-Bench metrics reproducible, we list the evaluation prompt templates used by the LLM planner and MLLM/VLM judges. These templates cover segment-importance planning, continuity and occurrence-group extraction, and video-prompt alignment judging. Placeholders such as \texttt{\{prompt\_lines\}}, \texttt{\{prompts\}}, and \texttt{\{num\_segments\}} are filled by the evaluator before sending the user message.

\begin{prompttemplatebox}{LLM Segment-Importance Planner Prompt}
You are the "segment-weight planner" for streaming video evaluation.

The input is a multi-segment prompt sequence for one video. The sequence
usually follows a beginning setup, middle development, and ending closure
structure. Based only on the prompts, assign each segment an integer
importance score from 1 to 100.

Please focus on:
1. Whether the segment carries a key transition in the before-after
   relationship. If so, assign a higher weight.
2. Whether the segment contains clear, verifiable, and non-substitutable
   action execution, character interaction, object transfer, state change,
   or task progress. If so, assign a higher weight.
3. Whether the segment is more likely to reveal whether the model truly
   understands and follows the prompt. If so, assign a higher weight.
4. Whether the segment is only repetition, setup, or closure. If so,
   assign a lower weight.
5. When semantic conditions are similar, prioritize event development and
   climax stages, respecting narrative structure.
6. The weights should be discriminative.

Requirements:
- The output length must match the number of input segments.
- Each score must be an integer from 1 to 100.
- Do not evaluate video quality; analyze only the prompts themselves.
- Output JSON only. Do not output explanations, Markdown, or any extra text.

Output format:
{"segment_importance":[int,int,int,int,int,int]}

Input prompt sequence:
{prompt_lines}
\end{prompttemplatebox}

\begin{prompttemplatebox}{CAC Continuity MLLM Judge Prompt}
Analyze the following sequence of video segment descriptions and determine
which adjacent segment pairs should preserve visual continuity, such as the
same scene, the same characters, or an obvious continuation.

Input segments:
{prompt_lines}

For each adjacent segment pair, output a Boolean judgment. The length must
be {num_transitions}.

Output format:
{"keep":[true,false,true]}
\end{prompttemplatebox}

\begin{prompttemplatebox}{CLC Occurrence-Group MLLM Extractor Prompt}
Analyze the following sequence of video segment descriptions. Identify the
main entities that appear, and mark the segment indices where each entity
appears (0-indexed).

{prompt_text}

Output JSON only, for example:
{"wizard":[0,1,2,3], "dragon":[1,2,3,4]}
\end{prompttemplatebox}

\begin{prompttemplatebox}{VLM Alignment Judge Prompt}
You are a video evaluation expert. Below is a streaming generated video and
its corresponding prompt sequence.
The system will provide key frames for each video segment in chronological
order. Judge whether each video segment executes the corresponding prompt,
and give an overall score for the entire sequence.

Prompt sequence:
{prompts}

Requirements:
1. `segment_scores` must be an array of length {num_segments}. The i-th
   element corresponds to the execution score of the i-th prompt segment.
2. All scores must be integers from 1 to 100. 1 means completely
   inconsistent, and 100 means completely consistent.
3. By default, score according to "the key frames of this segment" and
   "the prompt of this segment". Do not add points based on beginning,
   middle, or ending position, narrative structure, or segment importance.
4. If a segment prompt mainly describes continuation, maintenance,
   confirmation, observation, or still being in some state, you may use the
   state established in the previous segment to judge whether the state
   remains valid. Do not mechanically assign a low score only because the
   action magnitude in this segment is small.
5. Scoring must be based on directly observable evidence:
   - Whether the explicit action occurs
   - Whether character interaction occurs
   - Whether object change, object handoff, or state transition occurs
   - Whether the segment prompt is executed at the correct time
   - If the segment is a continuation/maintenance prompt, whether it clearly
     preserves the previously established character relationships, object
     states, and spatial layout without obvious conflicts
6. If only the characters, scene, or general atmosphere match, but the key
   action, interaction, object ownership, handoff receiving relation, or
   state change is unclear, `segment_score` should not exceed 50.
   For multi-person scenes, as long as character identity, interaction
   target, or object ownership is clearly uncertain, the score should not be
   high even if the overall image looks similar.
7. If there is almost no direct evidence supporting the segment prompt,
   `segment_score` should be in 1-20.
   If there is only weakly related, generic, or substitutable evidence,
   `segment_score` should usually be in 20-50, not high.
8. Adjacent segments may have the same score, and large jumps are allowed.
   Do not output an increasing or decreasing score sequence merely to make
   it look smooth. However, do not automatically assign an extremely low
   score to a "continuation holds" segment only because later-segment action
   is weaker.
9. `overall_score` indicates whether the whole video completes the key
   events in the correct temporal order and maintains later state
   consistency. If the key action, main interaction, object handoff, or
   state transition clearly occurs and later segments do not obviously
   violate the prompt, a medium-high score is allowed even if later segments
   have weaker motion. Conversely, if the video is only generally smooth,
   scene-consistent, and character-similar, but the key event, character
   relationship, or object ownership cannot be confirmed, `overall_score`
   should be clearly reduced.
10. Assign 100 only when character identity, key action, object relation,
    and timing are all very clear and almost unambiguous.

Output format:
{"segment_scores":[int,int,...],"overall_score":int}
\end{prompttemplatebox}

\paragraph{Planner and judge output schemas.}
For metric aggregation, the LLM planner sees only the six prompt texts and returns
\begin{equation}
\{\texttt{segment\_importance}: [w_1,\ldots,w_6]\},\quad w_i\in\{1,\ldots,100\}.
\end{equation}
The weights are normalized to sum to one. If the returned list has the wrong length, contains non-integer values, or cannot be parsed as JSON, the metric code falls back to uniform or metric-specific mean aggregation.

For VLM alignment in NarraStream-Bench, the judge receives the prompt sequence and the sampled video frames for each segment, then returns
\begin{equation}
\{\texttt{segment\_scores}: [q_1,\ldots,q_T],\texttt{overall\_score}: q\},
\quad q_i,q\in\{1,\ldots,100\}.
\end{equation}
Segment scores measure whether the visible evidence executes the corresponding prompt. The overall score measures whether the whole video completes the key events in order and preserves later state consistency. The parser normalizes these values to $[0,1]$ and falls back to the legacy scalar-score parser only for older cached responses.

\section{NarraStream-Bench Database Construction}
\label{app:narrstream_construction}

This appendix describes the construction of the NarraStream-Bench prompt database.
The benchmark is designed for narrative streaming generation: a model must keep a stable scene, follow evolving segment-level instructions, preserve entities after disappearance, and resolve partial or ambiguous references.

\paragraph{Scope and dimensions.}
Each sample is a 60-second single-scene narrative divided into six consecutive 10-second prompt segments.
We keep the scene identity fixed across the six segments so that failures are concentrated on streaming memory and instruction following rather than on full scene replacement.
The database is text-only: it contains prompt sequences and structured metadata, without scraped videos, personal data, or external visual assets.

Each prompt plan is labeled by six challenge dimensions:
\textit{initial character count} (\texttt{single}, \texttt{double}, \texttt{multi}),
\textit{interaction complexity} (\texttt{static}, \texttt{simple\_interaction}, \texttt{cooperative\_task}),
\textit{entry-exit dynamics} (\texttt{no\_entry\_exit}, \texttt{single\_entry\_exit}, \texttt{multi\_entry\_exit}),
\textit{temporal callback} (\texttt{no\_callback}, \texttt{adjacent\_callback}, \texttt{long\_range\_callback}),
\textit{character distinguishability} (\texttt{high}, \texttt{medium}, \texttt{low}), and
\textit{reference complexity} (\texttt{explicit\_description}, \texttt{partial\_reference}, \texttt{ambiguous\_reference}).
The selected combinations keep the benchmark broad enough to cover important marginal cases while avoiding mechanically combined scenarios that are visually unnatural or difficult to judge reliably.

\paragraph{Resources and sampling.}
The construction starts from three resources: a scene catalog, a blueprint catalog, and character-profile pools.
The scene catalog provides 9 stable venues with fixed \texttt{scene\_anchor} strings, concise background descriptions, and allowed pose modes.
The blueprint catalog provides 12 six-segment narrative templates, spanning control cases, common two-person interactions, multi-person cooperative tasks, entry-exit patterns, and long-range callbacks.
The profile pools provide names, stable visual attributes, partial references, and ambiguous references for different distinguishability levels.

The official suite contains 324 samples instantiated from 12 scene anchors, 9 blueprints, and 3 variants per scene-blueprint pair, for a total of 1,944 segment prompts.
Variants rotate character profiles and pose modes.
The pose mode (\texttt{seated}, \texttt{standing}, or \texttt{mixed}) is only a generation-diversity control, not an evaluation dimension.

\paragraph{Prompt plan construction.}
For every sampled scene-blueprint pair, we first render a structured prompt plan before asking an LLM to write natural-language prompts.
The plan records the scene, blueprint, six dimension labels, pose mode, segment beats, and a complete \texttt{character\_registry}.
The registry assigns each character a stable ID, role, display name, appearance description, explicit reference, partial reference, ambiguous reference, and first visible segment.
This registry provides the ground truth used by entity-grounding and cross-segment consistency metrics.

The plan also records three alignment structures.
The \texttt{presence\_matrix} specifies which character IDs should be visible in each segment, and \texttt{visible\_characters\_by\_segment} exposes this information to the prompt writer.
The \texttt{callback\_edges} field marks temporal dependencies, including adjacent callbacks and long-range callbacks.
The \texttt{reference\_mode\_by\_segment} field controls whether each segment uses explicit descriptions, partial references, or ambiguous references.

\paragraph{Prompt realization.}
The final six prompts are generated from the prompt plan using the templates in Appendix~\ref{app:agent_prompts}.
The first segment must introduce only the characters visible in segment 1 and use explicit visual descriptions.
Follow-up segments receive the previous prompt chain, current beat, visible and off-screen character lists, character registry, required reference mode, and pose guidance.
This setup keeps each segment self-contained enough for generation while still testing cross-segment memory.

\paragraph{Quality control and repair.}
After generation, each six-segment chain is validated by an automatic QC pass.
The validator checks the segment count, scene-anchor preservation, control-label leakage, non-English text, quotation marks or dialogue, prompt length, sentence count, unregistered character names, first-segment character introduction, and duplicate segment text.
Samples that fail QC are regenerated by resampling variant seeds under the same scene-blueprint pair, then rechecked.
The final NarraStream-Bench prompt suite contains 324 QC-passed samples and zero failed samples in the released metadata.

\section{Metric Details in NarraStream-Bench}
\label{app:narrstream_metrics}
\label{sec:supp_metric}

This appendix specifies the eleven NarraStream-Bench metrics. Each test sample contains
a prompt sequence $\{p_i\}_{i=1}^{T}$ and generated video segments $\{V_i\}_{i=1}^{T}$,
where $T=6$ in our benchmark. All metrics lie in $[0,1]$, with higher values indicating
better performance; tables in the main paper multiply them by $100$. Tunable
hyperparameters appear as symbols, with fixed values stated next to the corresponding
formula. Structural normalizations such as division by $255$, $2$, or $100$ remain
explicit. The same settings are used for all methods. Lowercase symbols denote
segment-, transition-, or entity-level scores, and uppercase $S$ with a
metric-specific subscript denotes the final sample-level score.

\paragraph{Narrative-aware Aggregation.}
NarraStream-Bench weights segment scores by narrative importance. Given the prompt
sequence, an LLM planner assigns integer importance values
$u_i\in\{1,\ldots,100\}$, which are normalized as
$w_i=u_i/\sum_{j=1}^{T}u_j$. We use
$\mathcal{A}_{\mathrm{seg}}$ for segment-level aggregation and
$\mathcal{A}_{\mathrm{tr}}$ for transition-level aggregation:
\[
\mathcal{A}_{\mathrm{seg}}(\{x_i\})=\sum_{i=1}^{T}w_i x_i,\qquad
\mathcal{A}_{\mathrm{tr}}(\{x_i\}_{i\in\Omega})=
\frac{\sum_{i\in\Omega}(w_i+w_{i+1})x_i}{\sum_{i\in\Omega}(w_i+w_{i+1})},
\]
where $\Omega$ is the set of valid transitions for the metric. If the planner response
is malformed or unavailable, we use the metric-specific fallback aggregation.
We use Qwen/Qwen3.5-27B as the LLM planner for segment-importance estimation.
The exact planner prompt is listed in Appendix~\ref{app:agent_prompts}.
Dataset-level metric scores average valid sample scores. Each group score is the
macro-average over its metrics, and Overall is the macro-average over the three groups:
\[
S_g=\frac{1}{|\mathcal{M}_g|}\sum_{m\in\mathcal{M}_g}S_m,\qquad
S_{\mathrm{overall}}=\frac{S_{\mathrm{quality}}+S_{\mathrm{temporal}}+S_{\mathrm{instruction}}}{3},
\]
where $\mathcal{M}_g$ is the metric set for group $g$: Visual Quality uses Subject Consistency (SC), Background Consistency (BC), Temporal Flickering (TF), Motion Smoothness (MS), and Video Temporal Stability Score (VTSS); Temporal Consistency uses Boundary Smoothness (BS), Conditional Adjacent Consistency (CAC), and Conditional Long-range Consistency (CLC); Instruction Compliance uses Entity Grounding (EG), Dynamic Trajectory (DT), and VLM Alignment (VLM).

\paragraph{Subject Consistency (SC) and Background Consistency (BC).}
SC measures subject appearance stability with DINO features~\cite{dino}, while BC measures scene
and background stability with CLIP features~\cite{clip}. For each metric
$m\in\{\mathrm{SC},\mathrm{BC}\}$, we sample five frames at relative positions
$\{0.1,0.3,0.5,0.7,0.9\}$ and extract normalized features
$\{\mathbf{x}_{i,k}^{m}\}$. Local consistency $\ell_i^{m}$ is the 25th percentile of
sampled frame-pair similarities within $V_i$. The feature medoid among sampled frames gives the
representative feature $\mathbf{r}_i^{m}$. Let
$c_{\mathrm{adj}}^{m}=\rho_{25}(\{\cos(\mathbf{r}_i^{m},\mathbf{r}_{i+1}^{m})\}_{i=1}^{T-1})$
denote adjacent-transition consistency and
$c_{\mathrm{anchor}}^{m}=\rho_{25}(\{\cos(\mathbf{r}_1^{m},\mathbf{r}_i^{m})\}_{i=2}^{T})$
denote first-segment-anchor consistency. We fuse them with local consistency as
\begin{equation}
S_m=\alpha_{\mathrm{cons}}\mathcal{A}_{\mathrm{seg}}(\{\ell_i^{m}\})
+(1-\alpha_{\mathrm{cons}})\cdot
\frac{c_{\mathrm{adj}}^{m}+c_{\mathrm{anchor}}^{m}}{2}.
\end{equation}
Here $\rho_{25}$ denotes the 25th percentile. We use
$\alpha_{\mathrm{cons}}=0.34$, so the cross-segment weight is $0.66$.

\paragraph{Temporal Flickering (TF).}
TF penalizes frame-to-frame brightness instability after motion compensation. For each
segment, we uniformly sample $K_{\mathrm{TF}}$ frames. For each adjacent pair
$(I_t,I_{t+1})$, RAFT~\cite{raft} estimates forward flow $F_{t\rightarrow t+1}$ and backward flow
$F_{t+1\rightarrow t}$. We warp the luminance $Y_{t+1}$ back to frame $t$ and compute
\begin{equation}
e_t(\mathbf{u})=\frac{|Y_t(\mathbf{u})-
\operatorname{warp}(Y_{t+1},F_{t\rightarrow t+1})(\mathbf{u})|}{255}.
\end{equation}
Residuals are measured only on in-bound pixels that pass a forward-backward flow
consistency check, which avoids treating occlusions or unreliable flow estimates as
flickering. We use $K_{\mathrm{TF}}=8$. If the valid mask covers less than $5\%$ of
pixels for a pair, we fall back to the in-bound pixels.
The pair error is the $\rho_{\mathrm{pair}}$ percentile of valid-pixel residuals. The
segment raw error $e_i$ is the $\rho_{\mathrm{seg}}$ percentile over pair errors, and the
segment score is
\begin{equation}
s_i^{\mathrm{TF}}=\exp(-e_i/\tau_{\mathrm{TF}}).
\end{equation}
We use $\rho_{\mathrm{pair}}=90$, $\rho_{\mathrm{seg}}=84$, and
$\tau_{\mathrm{TF}}=0.5$, where $\tau_{\mathrm{TF}}$ controls how quickly residual
error lowers the score. The final TF score is
$S_{\mathrm{TF}}=\mathcal{A}_{\mathrm{seg}}(\{s_i^{\mathrm{TF}}\})$.

\paragraph{Motion Smoothness (MS).}
MS evaluates whether the motion trajectory is temporally smooth. For each segment, an
AMT-S~\cite{amt} frame interpolation model predicts the midpoint frame between every second
frame pair. Let $\hat{I}_{2t+1}$ be the interpolated midpoint and $I_{2t+1}$ the
corresponding original frame. The raw interpolation error is
\begin{equation}
e_i=\frac{1}{N_i}\sum_t \operatorname{mean}_{\mathbf{u},c}
|I_{2t+1}(\mathbf{u},c)-\hat{I}_{2t+1}(\mathbf{u},c)| .
\end{equation}
It is mapped to a normalized segment score by
\begin{equation}
s_i^{\mathrm{MS}}=\exp(-e_i/\tau_{\mathrm{MS}}).
\end{equation}
We use $\tau_{\mathrm{MS}}=3.0$. The final MS score is
$S_{\mathrm{MS}}=\mathcal{A}_{\mathrm{seg}}(\{s_i^{\mathrm{MS}}\})$.

\paragraph{Video Temporal Stability Score (VTSS).}
VTSS uses a learned temporal-stability evaluator~\cite{koala36m,ivebench} to produce raw segment scores
$r_i$. We first compute the sample-level raw score
$r=\mathcal{A}_{\mathrm{seg}}(\{r_i\})$, then map it to $[0,1]$ with an anchored
linear transform:
\begin{equation}
S_{\mathrm{VTSS}}=
\operatorname{clip}\left(
\frac{r-r_{\mathrm{low}}}{r_{\mathrm{high}}-r_{\mathrm{low}}},0,1
\right).
\end{equation}
We use $r_{\mathrm{low}}=0.02$ and $r_{\mathrm{high}}=0.075$.

\paragraph{Boundary Smoothness (BS).}
BS measures whether the generated video changes smoothly at prompt boundaries. For the
transition between $V_i$ and $V_{i+1}$, we take the last two frames of $V_i$ and the
first two frames of $V_{i+1}$. RAFT~\cite{raft} gives the mean flow magnitude before the boundary
$m_i^{-}$, across the boundary $m_i^{b}$, and after the boundary $m_i^{+}$. The expected
boundary motion is $\bar{m}_i=(m_i^{-}+m_i^{+})/2$, and the normalized transition score is
\begin{equation}
s_i^{\mathrm{BS}}=\exp\left(
-\frac{|m_i^{b}-\bar{m}_i|}
{\epsilon_{\mathrm{BS}}+\lambda_{\mathrm{BS}}\bar{m}_i}
\right).
\end{equation}
We use $\epsilon_{\mathrm{BS}}=0.02$ and $\lambda_{\mathrm{BS}}=0.5$.
The final BS score is $S_{\mathrm{BS}}=\mathcal{A}_{\mathrm{tr}}(\{s_i^{\mathrm{BS}}\}_{i=1}^{T-1})$.

\paragraph{Conditional Adjacent Consistency (CAC).}
CAC evaluates adjacent segments only when the prompts imply that visual continuity
should be preserved. An MLLM judge reads $\{p_i\}$ and returns a Boolean keep flag for
each adjacent prompt pair. For each selected segment, we reuse the sampled DINO~\cite{dino} frame
features and define $\mathbf{v}_i$ as their feature medoid, yielding a single
representative segment vector. The raw cosine similarity is
$c_i=\cos(\mathbf{v}_i,\mathbf{v}_{i+1})$, which is remapped by
\begin{equation}
s_i^{\mathrm{CAC}}=
\operatorname{clip}\left(
\frac{c_i-\delta_{\mathrm{CAC}}}{1-\delta_{\mathrm{CAC}}},0,1
\right)^{\kappa_{\mathrm{CAC}}}.
\end{equation}
We use $\delta_{\mathrm{CAC}}=0.79$ and $\kappa_{\mathrm{CAC}}=2.0$.
The final CAC score is $S_{\mathrm{CAC}}=\mathcal{A}_{\mathrm{tr}}(\{s_i^{\mathrm{CAC}}\}_{i\in\Omega_{\mathrm{CAC}}})$,
where $\Omega_{\mathrm{CAC}}$ contains the selected continuity transitions.
We use Qwen/Qwen3-VL-30B-A3B-Instruct as the MLLM judge for selecting
continuity-preserving transitions. The exact judge prompt is listed in
Appendix~\ref{app:agent_prompts}.

\paragraph{Conditional Long-range Consistency (CLC).}
CLC measures whether an entity remains recognizable when it disappears and later
reappears. An MLLM extracts entity occurrence groups
$\mathcal{G}_e=\{i_1,\ldots,i_m\}$ from the prompt sequence. For each entity with at
least two occurrences, the first occurrence $i_1$ is the anchor. For each later
occurrence $i_j$, we compute its DINO~\cite{dino} similarity $c_{e,j}$ to the anchor and remap it as
\begin{equation}
s_{e,j}^{\mathrm{CLC}}=
\operatorname{clip}\left(
\frac{c_{e,j}-\delta_{\mathrm{CLC}}}{1-\delta_{\mathrm{CLC}}},0,1
\right)^{\kappa_{\mathrm{CLC}}}.
\end{equation}
We use $\delta_{\mathrm{CLC}}=0.58$ and $\kappa_{\mathrm{CLC}}=1.5$.
To aggregate reappearances of the same entity, we normalize the planner weights over target
occurrences $\{i_2,\ldots,i_m\}$ and lightly shrink them toward uniform weights. We use
$\lambda_{\mathrm{CLC}}=0.7$ for the planner-weighted component and $0.3$ for the uniform
component, and denote the resulting target-occurrence weight by $\tilde{w}_{e,j}$.
Let $\mathcal{E}^{+}$ be the valid reappearing entities. The entity and sample scores are
\begin{equation}
a_e=\sum_{j=2}^{m}\tilde{w}_{e,j}s_{e,j}^{\mathrm{CLC}},\qquad
S_{\mathrm{CLC}}=\frac{1}{|\mathcal{E}^{+}|}\sum_{e\in\mathcal{E}^{+}}a_e .
\end{equation}
Samples with no valid reappearance group are omitted from the CLC dataset average.
We use Qwen/Qwen3-VL-30B-A3B-Instruct as the MLLM judge for extracting
long-range entity occurrence groups. The exact extractor prompt is listed in
Appendix~\ref{app:agent_prompts}.

\paragraph{Entity Grounding (EG).}
EG checks whether prompt-specified entities and their visible attributes are grounded in
the generated segment. For each prompt $p_i$, an LLM extracts up to four visually
verifiable entities and up to five visible attributes per entity. A VLM receives five
sampled frames from $V_i$ and returns, for each entity $e$, a presence score $a_{i,e}$
and an attribute-match score $m_{i,e}$ in $[0,1]$. The entity score is
$g_{i,e}=a_{i,e}m_{i,e}$, and the segment score is
\begin{equation}
s_i^{\mathrm{EG}}=
\begin{cases}
\frac{1}{|\mathcal{E}_i|}\sum_{e\in\mathcal{E}_i}g_{i,e}, & |\mathcal{E}_i|>0,\\
\nu_{\mathrm{EG}}, & |\mathcal{E}_i|=0.
\end{cases}
\end{equation}
We use $\nu_{\mathrm{EG}}=0.5$. The final EG score is
$S_{\mathrm{EG}}=\mathcal{A}_{\mathrm{seg}}(\{s_i^{\mathrm{EG}}\})$.
We use Qwen/Qwen3.5-27B for prompt-side entity extraction and
Qwen/Qwen3-VL-30B-A3B-Instruct for frame-based visual scoring.

\paragraph{Dynamic Trajectory (DT).}
DT rewards visual change when the prompt changes substantially and visual stability when
the prompt is mostly continuous. LanguageBind~\cite{languagebind} encodes each prompt and video segment.
For transition $i$, let $d_i^{p}=1-\cos(\mathbf{t}_i,\mathbf{t}_{i+1})$ and
$d_i^{v}=1-\cos(\mathbf{z}_i,\mathbf{z}_{i+1})$, where $\mathbf{t}$ and
$\mathbf{z}$ are normalized text and video features. A soft gate
determines whether the transition should favor response or stability:
\begin{equation}
\gamma_i=\sigma\left(\frac{d_i^{p}-\mu_{\mathrm{DT}}}{\tau_{\mathrm{gate}}}\right).
\end{equation}
Here $\sigma(x)=1/(1+\exp(-x))$ is the logistic sigmoid.
We use $\mu_{\mathrm{DT}}=0.25$ and $\tau_{\mathrm{gate}}=0.05$.
The response term is $r_i=1-\exp(-d_i^{v}/\tau_{\mathrm{on}})$, and the stability
term is $h_i=\exp(-d_i^{v}/\tau_{\mathrm{off}})$. We use
$\tau_{\mathrm{on}}=0.02$ and $\tau_{\mathrm{off}}=0.06$. The transition score is
\begin{equation}
s_i^{\mathrm{DT}}=\gamma_i r_i+(1-\gamma_i)h_i.
\end{equation}
The final DT score is $S_{\mathrm{DT}}=\mathcal{A}_{\mathrm{tr}}(\{s_i^{\mathrm{DT}}\}_{i=1}^{T-1})$.

\paragraph{VLM Alignment (VLM).}
VLM Alignment directly evaluates prompt-following with a multimodal judge. The judge
receives the complete prompt sequence and five uniformly sampled frames from each
segment, then returns per-segment integer scores $q_i$ and an overall integer score
$q$, all in $\{1,\ldots,100\}$.
After normalization to $[0,1]$, the segment component is
$Q_{\mathrm{seg}}=\mathcal{A}_{\mathrm{seg}}(\{q_i/100\})$. The final VLM score combines
segment-level prompt execution and whole-video order/state consistency:
\begin{equation}
S_{\mathrm{VLM}}=\alpha_{\mathrm{VLM}}Q_{\mathrm{seg}}+(1-\alpha_{\mathrm{VLM}})(q/100).
\end{equation}
We use $\alpha_{\mathrm{VLM}}=0.8$, so the overall-score weight is $0.2$.
We use Qwen/Qwen3-VL-30B-A3B-Instruct as the multimodal judge for VLM Alignment.
The exact judge prompt and output schema are listed in Appendix~\ref{app:agent_prompts}.

\section{Baseline Implementation Details}
\label{app:baseline_implementation}

The official implementations of Self Forcing~\cite{selfforcing}, Rolling Forcing~\cite{rollingforcing}, and Deep Forcing~\cite{deepforcing} do not support interactive multi-prompt generation, where prompts change during an ongoing autoregressive rollout.
For NarraStream-Bench, we add a common prompt-switching wrapper to these baselines.
Each method follows the same six-segment schedule as IAMFlow, generates a 60-second video at $832{\times}480$ resolution, and switches prompts at segment boundaries while preserving the generated history.

We do not switch prompts by replacing only the text-encoder tokens, because the temporal KV cache states would remain conditioned on the previous prompt and weaken the baselines.
Following LongLive~\cite{longlive}, we instead refresh the active causal context with KV-recache at each prompt switch: after encoding the new prompt, we rerun the DiT on the context window used by each method under the new text condition before continuing generation.
We keep each baseline's original long-video mechanism unchanged, including Self Forcing's autoregressive rollout model, Rolling Forcing's rolling context update, and Deep Forcing's deep-sink and participative-compression strategy.
For the efficiency comparison, we use the same two-GPU deployment budget for baselines and offload VAE decoding to the second GPU when applicable.

\section{Human Alignment Study Details}
\label{app:human_study}

\begin{figure}[t]
  \centering
  \includegraphics[width=\linewidth]{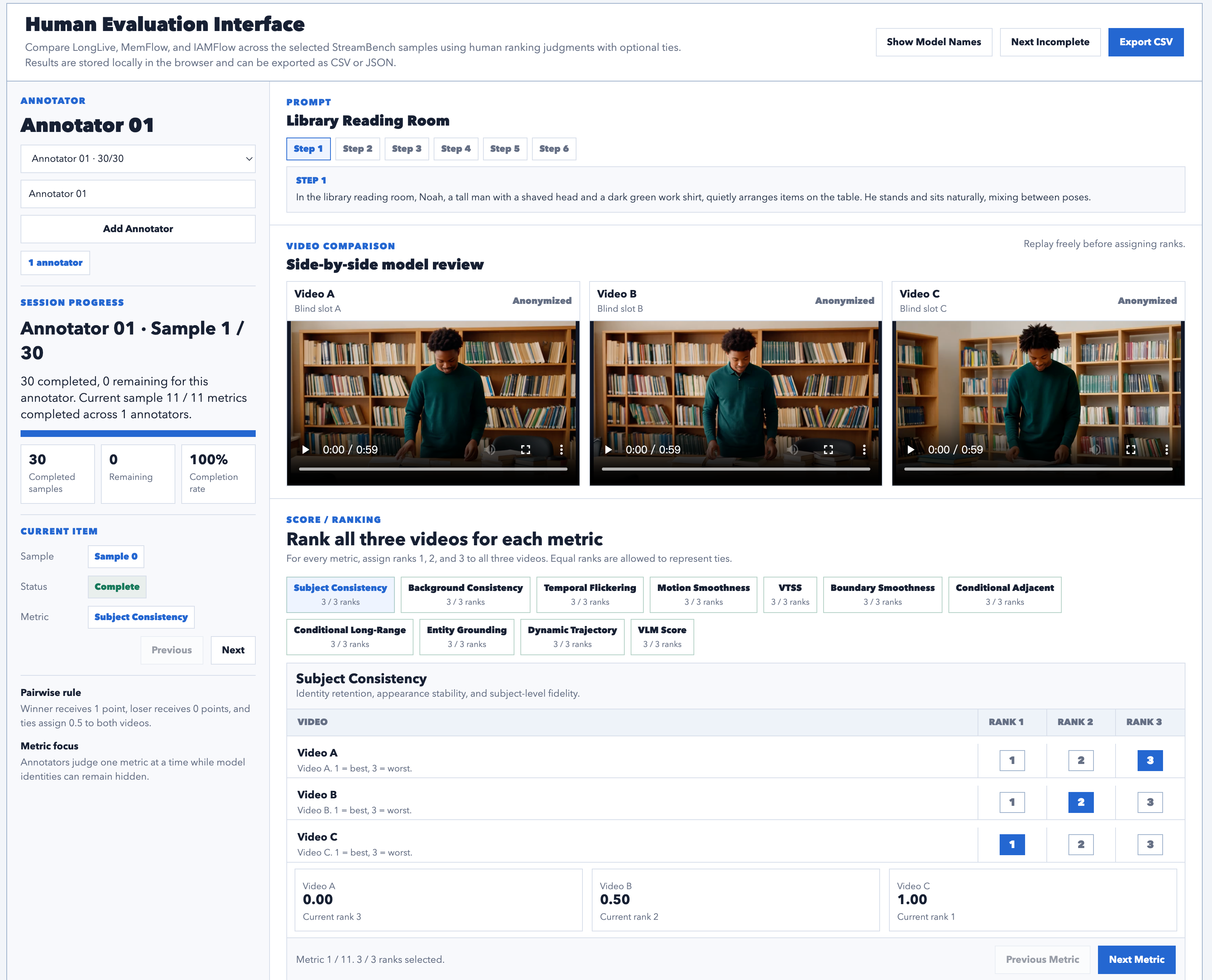}
  \caption{
  \textbf{Human alignment study interface}.
  Participants view the full prompt sequence and compare anonymized videos before ranking each metric.}
  \label{fig:app_human_alignment}
\end{figure}

We conduct the human alignment study to test whether the automatic metrics in
NarraStream-Bench agree with human judgments on model ordering. We sample 30
benchmark cases and compare the same three methods as the main comparison: LongLive,
MemFlow, and IAMFlow. Each case contains the six-prompt narrative sequence and three
60-second generated videos. We recruit 30 volunteers, and each volunteer evaluates all
cases independently. The annotation interface is shown in Fig.~\ref{fig:app_human_alignment}.

\paragraph{Annotation protocol.}
Participants view the full prompt sequence above the generated videos so that they can
judge the full set of NarraStream-Bench metrics. The study is blind:
model names are hidden and videos are shown as \textit{Video A}, \textit{Video B},
and \textit{Video C}, with the display order shuffled per sample. Participants
evaluate the same 11 dimensions defined in Appendix~\ref{app:narrstream_metrics},
covering video quality, temporal consistency, and instruction compliance. For each
sample and metric, they rank the three videos from 1 to 3, where rank 1 indicates the
best result. Ties are allowed when videos are visually indistinguishable for the
current metric.

\paragraph{Pairwise preference aggregation.}
We convert ranks into pairwise preferences to match the scoring protocol in the main
text. For annotator $k$, sample $i$, metric $m$, and methods $a$ and $b$ with ranks
$r_{k,i,m,a}$ and $r_{k,i,m,b}$, the induced pairwise score is
\begin{equation}
s_{k,i,m}(a,b)=
\begin{cases}
1, & r_{k,i,m,a}<r_{k,i,m,b},\\
0, & r_{k,i,m,a}>r_{k,i,m,b},\\
0.5, & r_{k,i,m,a}=r_{k,i,m,b}.
\end{cases}
\end{equation}
We first average pairwise scores across annotators for the same sample, metric,
and ordered model pair:
\begin{equation}
\bar{s}_{i,m}(a,b)=
\frac{1}{|\mathcal{K}_{i,m}|}
\sum_{k\in\mathcal{K}_{i,m}} s_{k,i,m}(a,b),
\end{equation}
where $\mathcal{K}_{i,m}$ denotes annotators with a valid annotation for metric $m$
on sample $i$. This produces one human preference value for each sample, metric,
and model pair.

\paragraph{Alignment with automatic metrics.}
Let $A_{i,m}(a)$ be the automatic NarraStream-Bench score of method $a$ on sample
$i$ and metric $m$. For the same ordered pair $(a,b)$, we compute the automatic
pairwise margin as
\begin{equation}
\Delta A_{i,m}(a,b)=A_{i,m}(a)-A_{i,m}(b).
\end{equation}
For each metric, we compute Spearman's rank correlation over all samples and all
three unordered model pairs:
\begin{equation}
\rho_m=\operatorname{Spearman}\left(
\{\bar{s}_{i,m}(a,b)\}_{i,(a,b)},
\{\Delta A_{i,m}(a,b)\}_{i,(a,b)}
\right),
\end{equation}
where $i$ ranges over the 30 selected samples and $(a,b)$ denotes one fixed
ordering of each of the three unordered model pairs drawn from
$\mathcal{M}=\{\text{LongLive},\text{MemFlow},\text{IAMFlow}\}$.
Thus each metric is evaluated with 90 pairwise comparison points. Higher $\rho_m$
means that the automatic metric gives pairwise preferences more similar to human
annotators.
Table~\ref{tab:tb7_human_alignment} reports the resulting correlation for each metric.

\section{Limitation Analysis}
\label{app:limitations}
\begin{figure}[t]
  \centering
  \includegraphics[width=\linewidth]{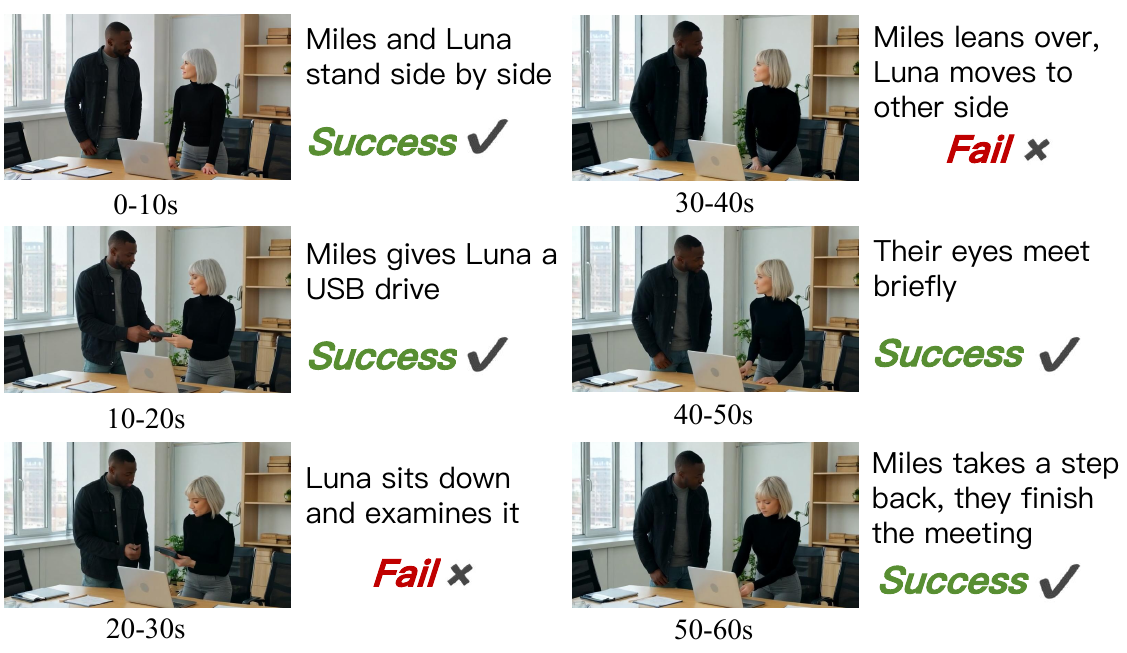}
  \caption{\textbf{Failure case of IAMFlow}. Our method preserves entity identities, but struggles with complex instructions and dynamic actions.}
  \label{fig:failure_case}
\end{figure}

We inspect challenging blueprints with low distinguishability, ambiguous references, and rapidly evolving instructions. Although our method improves long-range consistency through entity- and attribute-centric memory, this design can become overly conservative when the prompt requires fast adaptation to dynamic scene changes or semantically complex edits. In particular, when an entity suddenly changes roles, appearance, or interactions, the retrieved memory may still over-emphasize previously verified identity cues, causing the generator to lag behind the newest instruction.
As shown in Fig.~\ref{fig:failure_case}, IAMFlow often preserves the coarse identity of the main subject, but may miss fine-grained semantic updates such as transient state changes, relational constraints, or newly introduced compositional actions. This tension reflects a trade-off between stable identity preservation and flexible semantic control. Balancing persistent entity memory, prompt responsiveness, and cross-segment coherence remains an open challenge for streaming long-video generation.

\section{More Qualitative Results and Details}
\label{app:more_qual_details}

\begin{figure}[t]
  \centering
  \includegraphics[width=\linewidth,height=0.68\textheight,keepaspectratio]{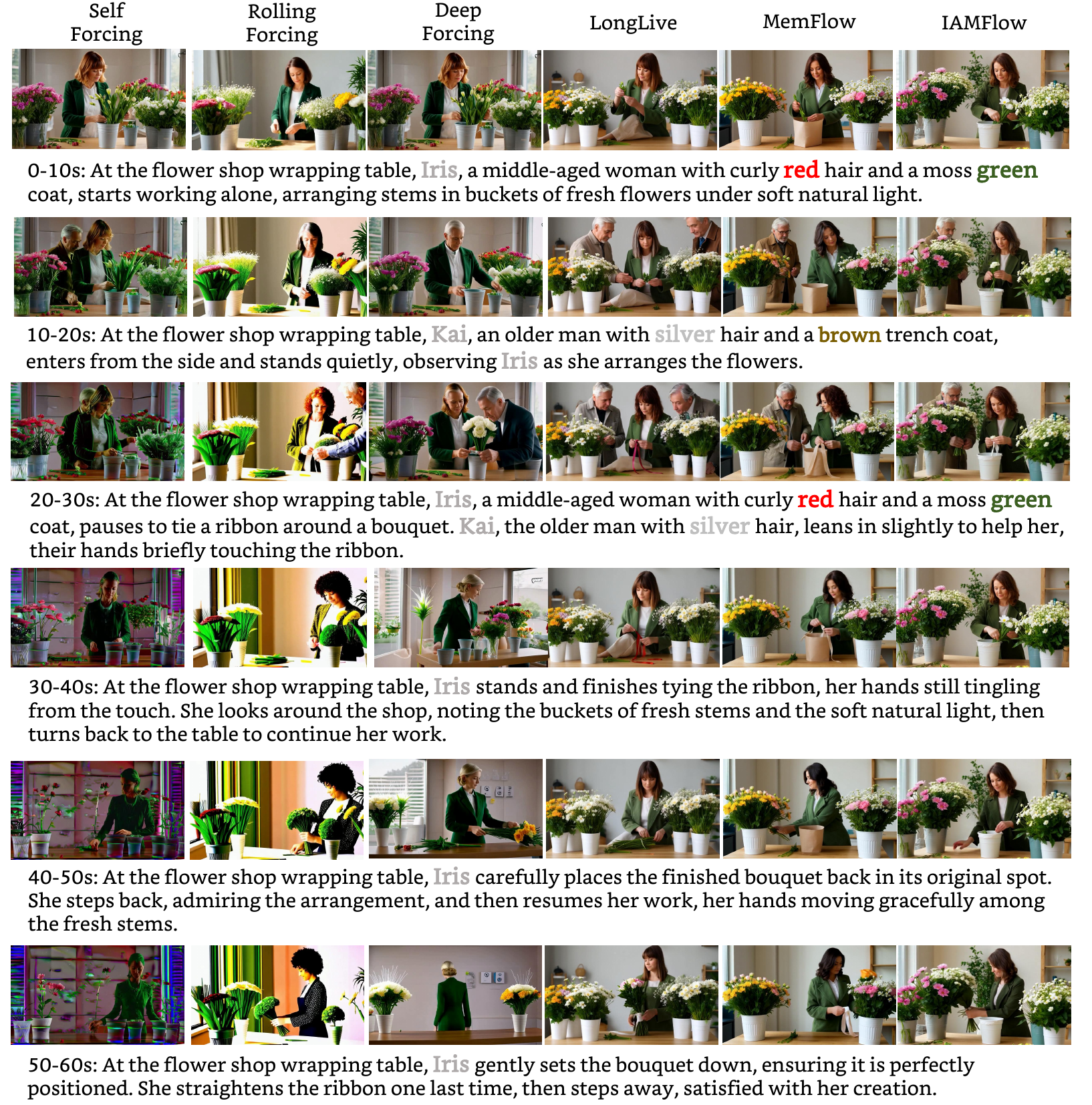}
  \caption{
  \textbf{Two-person interaction scenarios.}
  IAMFlow preserves the number of entities and their visual attributes across long videos, showing reliable identity-aware memory under evolving two-person interactions.}
  \label{fig:app_more_qualitative_1}
\end{figure}

\begin{figure}[t]
  \centering
  \includegraphics[width=\linewidth,height=0.78\textheight,keepaspectratio]{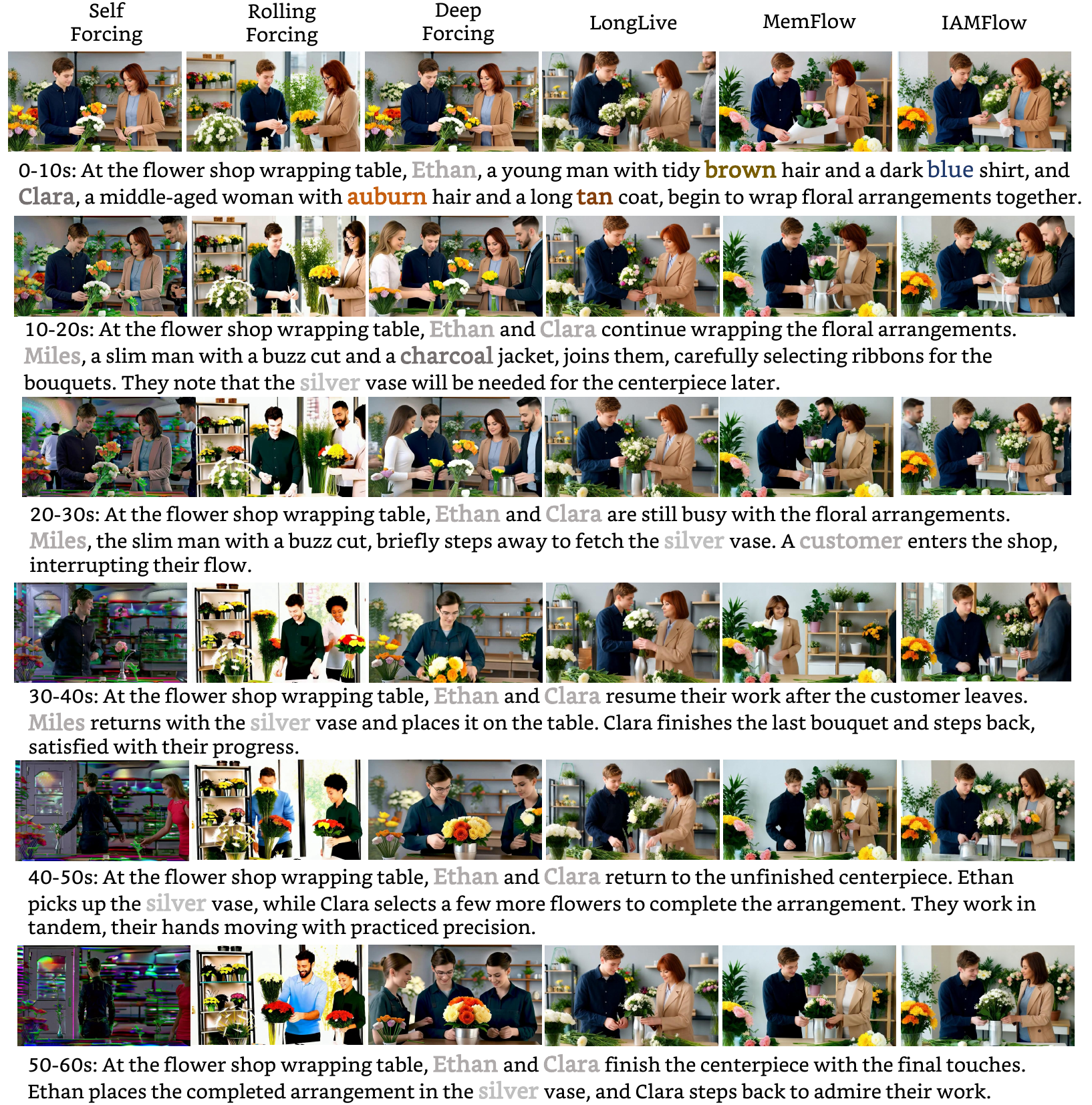}
  \caption{
  \textbf{Multi-person interaction scenarios.}
  IAMFlow maintains strong long-term consistency and instruction following while keeping multiple interacting entities stable throughout the generation.}
  \label{fig:app_more_qualitative_2}
\end{figure}

\begin{figure}[t]
  \centering
  \includegraphics[width=\linewidth,height=0.78\textheight,keepaspectratio]{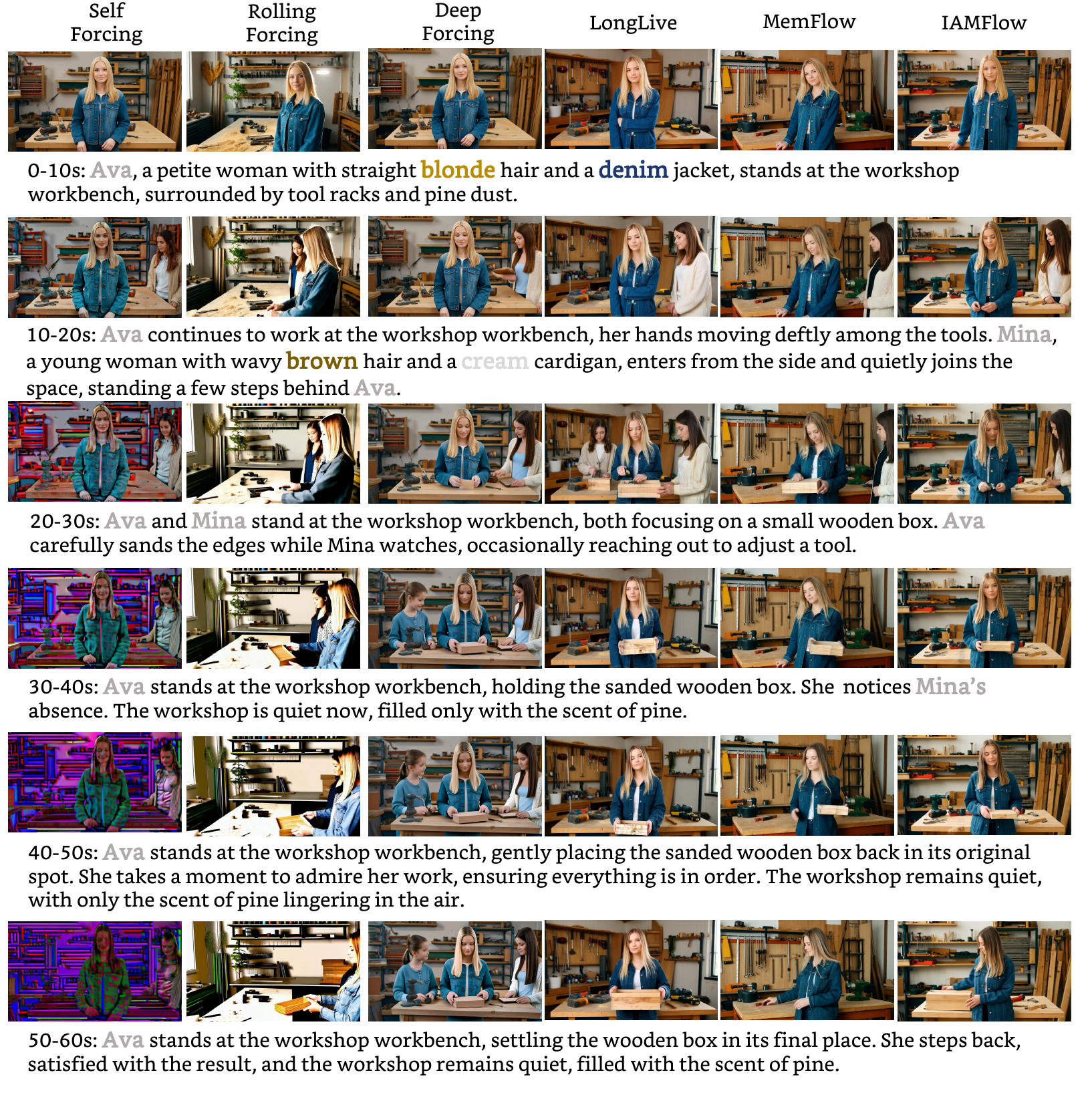}
  \caption{
  \textbf{Two-person interactions across diverse scenes.}
  IAMFlow generalizes to different scene settings while preserving stable entity identities, attributes, and visual composition across long videos.}
  \label{fig:app_more_qualitative_3}
\end{figure}

Additional qualitative results of our method are shown in Fig.~\ref{fig:app_more_qualitative_1}, Fig.~\ref{fig:app_more_qualitative_2}, and Fig.~\ref{fig:app_more_qualitative_3}. These examples show that the training-free IAMFlow framework improves long-video identity preservation, temporal consistency, and instruction compliance, producing results on par with or better than training-based baselines.

\clearpage
\section{Ethics Statement}
\label{app:ethics}
IAMFlow is a training-free inference framework and does not introduce additional external video datasets. 
NarraStream-Bench consists of text-only narrative prompts generated with Qwen2.5-7B-Instruct and Qwen2.5-72B-Instruct. The prompts are intended for academic evaluation and contain no external video data. 

\section{Reproducibility Statement}
\label{app:reproduce}
To support reproducibility, we will open-source this project, including the inference code and quantized model weights, as well as the database and evaluation scripts of NarraStream-Bench. 
We provide the full implementation details of our method and benchmark in Sec.~\ref{sec:method} and Appendices~\ref{app:memory_implement}, \ref{app:narrstream_construction}, and \ref{app:narrstream_metrics}.

\section{Broader Social Impact}
\label{app:social}
Efficient narrative video generation can lower the cost of long-form visual creation. 
This may benefit education, storyboarding, accessibility tools, creative prototyping, small studios, and researchers without access to expensive video production pipelines. 
Since IAMFlow improves long-range consistency without additional video-model training, it may also reduce the compute needed to study interactive long-video generation.

However, the same capabilities can be misused. More coherent long videos may make fabricated events, synthetic identities, or misleading visual narratives harder for viewers to detect. 
Identity-aware memory can also preserve biased or stereotyped visual attributes if the base generator, LLM, VLM, or benchmark prompts encode such biases. 
In deployment settings, these risks should be addressed through content provenance, visible labeling, digital watermarking, abuse monitoring, safety filters, and policies that forbid non-consensual likeness generation and deceptive use. 
We also encourage continued collaboration among researchers, platform operators, policymakers, and civil-society groups on transparent data practices and evaluation protocols for synthetic video.

\end{document}